\documentclass{bmvc2k}

\title{Self-Supervised Point Cloud Completion \\ via Inpainting
}

\addauthor{Himangi Mittal}{hmittal@andrew.cmu.edu}{1}
\addauthor{Brian Okorn}{bokorn@andrew.cmu.edu}{1}
\addauthor{Arpit Jangid}{ajangid@andrew.cmu.edu}{1}
\addauthor{David Held}{dheld@andrew.cmu.edu}{1}

\addinstitution{
 Carnegie Mellon University\\
 Pittsburgh, PA, USA
}

\runninghead{Mittal, okorn, jangid, held}{Self-Supervised Point Cloud Completion}

\def\etal{\emph{et al}\bmvaOneDot}

\usepackage{epsfig}
\usepackage{graphicx}
\usepackage{amsmath}
\usepackage{amssymb}
\usepackage{multirow}
\usepackage{wrapfig}
\usepackage{bbm}
\usepackage{xcolor} 

\usepackage{color, soul} 
\usepackage{multicol}
\begin{document}

\maketitle

\maketitle

\begin{abstract}
When navigating in urban environments, many of the objects that need to be tracked and avoided are heavily occluded.  Planning and tracking using these partial scans can be challenging. The aim of this work is to learn to complete these partial point clouds, giving us a full understanding of the object's geometry using only partial observations. Previous methods achieve this with the help of complete, ground-truth annotations of the target objects, which are available only for simulated datasets. However, such ground truth is unavailable for real-world LiDAR data. In this work, we present a self-supervised point cloud completion algorithm, PointPnCNet, which is trained only on partial scans without assuming access to complete, ground-truth annotations. Our method achieves this via inpainting. We remove a portion of the input data and train the network to complete the missing region. As it is difficult to determine which regions were occluded in the initial cloud and which were synthetically removed, our network learns to complete the full cloud, including the missing regions in the initial partial cloud. We show that our method outperforms previous unsupervised and weakly-supervised methods on both the synthetic dataset, ShapeNet, and real-world LiDAR dataset, Semantic KITTI.
\end{abstract}
\begin{figure}[t]
\centering
\includegraphics[trim=0 0 0 0, clip, width=0.8\columnwidth]{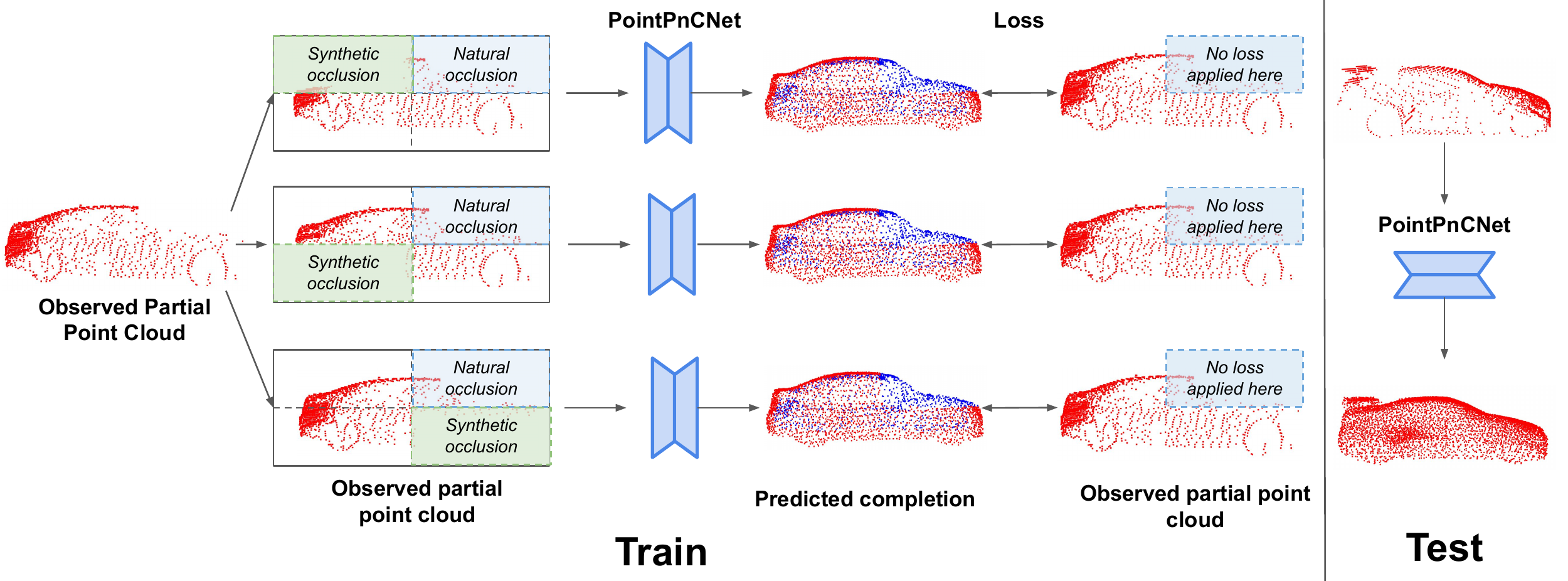}
\vspace{-1em}
 	\caption{We adopt an \textit{inpainting}-based approach for self-supervised point cloud completion to train our network using only partial point clouds. Given a partial point cloud as input, we randomly remove regions from it and train the network to complete these regions using the input as the pseudo-ground truth. The loss is only applied to the regions which have points in the observed input partial point cloud (\textcolor{red}{red}). Since, the network cannot differentiate between synthetic and natural occlusions, the network predicts a complete point cloud.}
\label{fig:first_page_image}
\vspace{-1.5em}
\end{figure}



\vspace{-1.5em}
\section{Introduction}
\vspace{-0.5em}

Autonomous vehicles often understand the world around them using depth sensors such as LiDAR. However, the LiDAR point clouds are often incomplete even when recorded from multiple viewpoints over time.
To accurately track objects and plan routes to avoid collisions, it is important for autonomous vehicles to understand the complete shape of surrounding objects.


Previous methods~\cite{yuan2018pcn, tchapmi2019topnet, wang2020cascaded, xie2020grnet, wen2020point} have learned to complete partial point clouds but they strongly rely on the availability of ground truth complete shapes as supervision. Since complete point clouds are costly to obtain for real-world scenarios, these methods typically train only from simulated data where ground-truth completions are available. This limitation motivates our approach to learn only from partial point clouds to complete shapes without ever observing the ground-truth completed point clouds during training.

To this end, our method leverages self-supervision via an inpainting-based approach where we randomly remove regions from the partial point clouds and train the network to complete the entire point cloud. Across multiple training examples, different regions will be occluded and varying regions will be synthetically removed. Because the network does not know which regions were artificially removed and which were naturally occluded in each original partial point cloud, the network learns to attempt to complete the entire point cloud. 

In contrast to images, where a mask can specify the region to inpaint~\cite{liu2020rethinking, pathakCVPR16context, yu2018generative, yeh2017semantic, hong2019deep}, the unstructured nature of point clouds makes it challenging to define which regions the network needs to inpaint. We solve this using a region-aware loss which penalizes only the regions where the original point cloud was present. Additionally, we partition the point cloud into local regions using intersecting half-spaces and encode/decode them separately. This allows the network to learn data-driven embeddings separately for each local region that specialize in individual region-level object parts.  We also encode/decode at the global point cloud level to further allow the network focus on each region jointly with each other. Previous works \cite{insafutdinov2018unsupervised, gu2020weakly} depend on aligning multiple viewpoints of an object during training which can be sensitive to pose alignment errors. While we also incorporate a multi-viewpoint loss, we show that our use of inpainting allows our method to be robust to alignment errors. 

The key contributions of this paper are as follows:  1).  We present a novel inpainting-based self-supervised algorithm that learns to complete missing local regions in an incomplete point cloud without the need for ground truth point cloud completions, 2).  Our multi-level encoder-decoder based architecture, PointPnCNet, partitions the point clouds to learn local and global embeddings to obtain improved completion performance, 3).Our approach outperforms existing methods for unsupervised point cloud completion~\cite{insafutdinov2018unsupervised, gu2020weakly} when evaluated on the standard completion benchmarks of ShapeNet~\cite{chang2015shapenet} and SemanticKITTI~~\cite{behley2019semantickitti}.

\vspace{-1.25em}
\section{Related Work}
\vspace{-0.5em}


\textbf{Supervised Point Cloud Completion} Most of the existing 3D shape completion methods~\cite{yuan2018pcn, tchapmi2019topnet, wang2020cascaded, xie2020grnet, wen2020point, wang2020softpoolnet, huang2020pf} make use of complete ground-truth shape labels. A common approach for point cloud completion is to use an encoder-decoder style architecture~\cite{yuan2018pcn,liu2020morphing,xie2020grnet,wang2020point}. On the other hand, Tchapmi \etal~\cite{tchapmi2019topnet} proposed to generate a point cloud using a hierarchical rooted tree structure. Our architecture builds on the typical encoder-decoder style of previous work~\cite{yuan2018pcn}. In contrast to the above supervised methods, our proposed approach does not require ground truth annotations. This allows our method to be trained using LiDAR data in the wild, as opposed to the previous methods which are trained only with simulated data.

\textbf{Weakly-Supervised Methods}
Recently, Gu~\etal~\cite{gu2020weakly} proposed a weakly-supervised approach for point cloud completion where the pose of the input partial point cloud and 3D canonical shape are jointly optimized. Their method is weakly-supervised via multi-view consistency among the multiple partial observations of the same instance. Our method also uses partial point clouds, however, using our inpainting-based approach, our method is able to learn a more accurate completion and is robust to view alignment errors. Other methods also learn 3D shape reconstruction using weak supervision~\cite{yan2016perspective, zhu2017rethinking, tulsiani2018multi}. Among these, Differentiable Point Clouds~(DPC)~\cite{insafutdinov2018unsupervised} jointly predicts camera poses and a 3D shape representation given two image views of the same instance. The geometric consistency between the estimated 3D shape and the input images is enforced using an end-to-end differentiable point cloud projection.  We show in the results that we significantly outperform this method.



\textbf{Image Inpainting}
In the area of image inpainting~\cite{liu2020rethinking, pathakCVPR16context, yu2018generative, yeh2017semantic, hong2019deep, zhan2020self}, Zhan \etal~\cite{zhan2020self} proposed self-supervised partial completion networks (PCNets) to complete an occluded object's mask and content in the input image. Our method takes inspiration from Zhan \etal~\cite{zhan2020self} for shape completion in 3D point clouds. 
However, due to the structured nature of images, often a mask can be used to specify the region to inpaint. In 3D point clouds, where the data is unstructured and sparse in nature, it is difficult to specify a ``mask" for the regions to inpaint. There is ambiguity between regions that have been ``masked out" and regions that are naturally occluded, making the task of inpainting challenging for point cloud data.

\textbf{Point Cloud Inpainting} Some of the previous methods~\cite{fu2018point, yu2020point, fu20203d, zhao2020pui, chen2020point, hu2019local} have also explored inpainting in the point cloud domain. However, these methods either use ground truth during training~\cite{yu2020point, zhao2020pui}, rely on template-matching within a data sample~\cite{fu2018point, fu20203d, hu2019local}, or project a point cloud into 2D structured representation~\cite{chen2020point}. Our method is novel in the sense that it uses inpainting directly on the point clouds without any ground truth information while leveraging large datasets to learn domain-specific priors. 

\vspace{-1.25em}
\section{Method}
\vspace{-0.5em}

The point cloud completion problem can be defined as follows: given an incomplete set of sparse 3D points $X$, sampled from a partial view of an underlying dense object geometry $G$, the goal is to predict a new set of points $Y$, which mimics a uniform sampling of $G$. 


\begin{figure*}[t]
 	\centering
 	\includegraphics[trim=0 20 0 0, clip, width=1.0\textwidth]{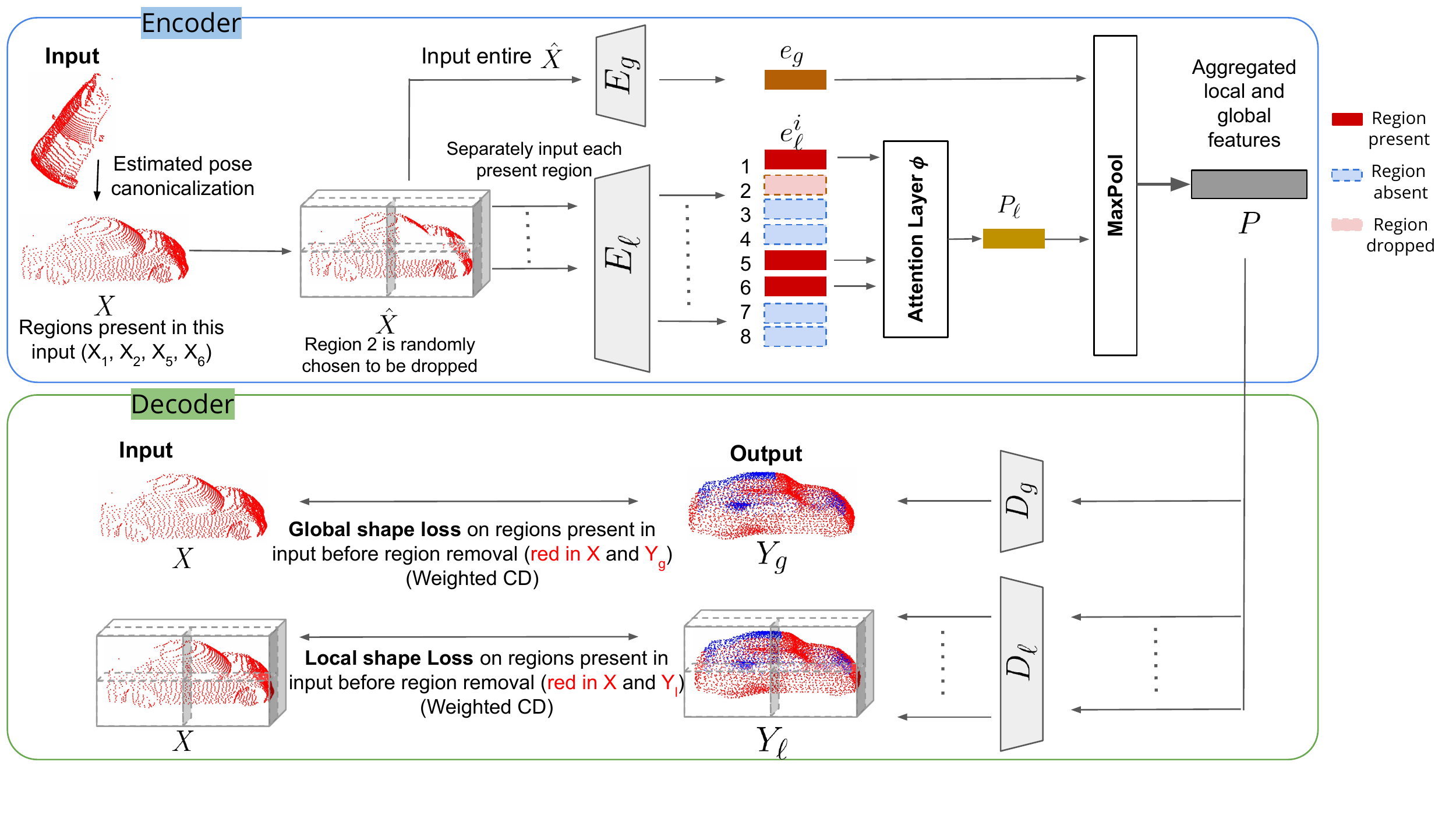}
 	\vspace{-2em}
 	\caption{\textbf{PointPnCNet Architecture:} Our method first estimates a canonicalized orientation of a partial point cloud, which has some regions missing due to natural occlusions.  We then randomly drop one or more of the regions to create additional synthetic occlusions. We compute global features $e_g$ and local features $P_\ell$ which we combine into an encoding $P$. Our multi-level decoder uses the encoding $P$ to generate a completed point cloud. The global shape loss and local shape loss are only applied to the regions of the output where points are present in the original cloud (before synthetic occlusions) which are shown in \textcolor{red}{red} in $X$, $Y_g$, and $Y_\ell$. The \textcolor{blue}{blue} points in $Y_g$ and $Y_\ell$ are not present in the original cloud, so we have no ground truth about their positions; thus they are not penalized in the loss. The final output of the network is the concatenation of the outputs from $Y_g$ and $Y_\ell$.}
 	\label{fig:model_diagram}
 	\vspace{-1.5em}
\end{figure*} 
\vspace{-1em}
\subsection{Self-supervised Inpainting}
\label{sec:inpainting}
\vspace{-0.5em}
In our self-supervised inpainting-based approach to learn to complete full point clouds using only partial point clouds, we randomly remove regions of points from a given partial point cloud and train the network to inpaint these synthetically removed regions. The original partial point cloud is then used as a pseudo-ground truth to supervise the completion. Since we do not have the complete ground-truth point cloud, supervision is only applied to the regions of the original point cloud that contain points (i.e. unoccluded regions). 
The network leverages the information of available regions across samples and embeds each region separately that can generalize across partially occluded samples with different missing regions. Further, due to the stochastic nature of region removal, the network cannot easily differentiate between the synthetic and original occlusions of the input partial point cloud, making the network learn to complete the point cloud. Thus, the combination of inpainting, random-region removal, and region-specific embeddings enables the model to generate all the regions and create a complete point cloud.

\vspace{-1em}
\subsection{Network Architecture}
\vspace{-0.5em}

Figure~\ref{fig:model_diagram} depicts the architecture and training flow of our network, Point Cloud Partition-and-Completion Network (PointPnCNet). We use a multi-level encoder-decoder architecture to allow the network to focus on different parts of an object. We present the evaluations of various alternate designs of our method in the appendix.
\vspace{-0.5em}
\subsubsection{Multi-Level Encoder}
\label{sec:multi-scale-encoder}
\vspace{-0.5em}
 Our encoder consists of multiple, parallel encoder streams that encode the input partial point cloud at global and local levels. The global-level encoder operates on the full-scope of the object while a local-level encoder focuses on a particular region of the object. Since a local encoder only sees points in a given local region and is invariant to other parts of the shape which may be missing, local encoders make the network robust to occlusions by focusing on individual object parts separately. Global encoder further enhances shape consistency by focusing on regions jointly with each other.
 Given a partial point cloud, we estimate its canonical frame using a learned method~(Sec.~\ref{sec:experiments}) and transform it to obtain a canonicalized partial point cloud $X$. We show that our method is robust to errors in this canonicalization~(Sec.~\ref{sec:robustness}).  
 We then partition the canonicalized partial point cloud using intersecting half-spaces that are produced by the coordinate planes after canonicalization. This effectively separates the space into eight 3D octants as shown in Figure~\ref{fig:model_diagram}. While other types of partitioning could be used, we found this subdivision to be simple and reasonably effective. Rather than a strict partitioning, we allow a small overlap between neighboring regions such that points in the overlap are present in both regions. This helps to avoid seams at boundaries.
Let $X_{i}$ consist of the points in the region $i$. After partitioning, we remove points of any particular region with a probability $p$ to simulate a synthetic occlusion. We use this synthetically occluded point cloud $\hat{X}$ as input to our inpainting network.
The points in the remaining regions are aggregated together and passed as input to the global encoder, $E_{g}$, to give a global embedding $e_g$~(Figure~\ref{fig:model_diagram}). 
In parallel, each remaining region $X_{i}$ is separately encoded by a local encoder, $E_{\ell}$ to obtain a local embedding for that region, $e^{i}_{\ell}$. To aggregate the local feature embeddings, each embedding $e^{i}_{\ell}$ is fed as input into an attention module, consisting of an MLP layer, that generates a set of weights $w_{i} = \phi (e^{i}_\ell)$. These weights are used to weigh each of the embeddings $e^{i}_{\ell}$ in a linear combination to form the aggregate embedding $P_\ell = \sum_{i} \mathbbm{1}_i w_{i} \, e^{i}_\ell$, where $\mathbbm{1}_i$ is an indicator function which equals 1 if region $i$ is present~(i.e. present in the original partial point cloud $X$ and not randomly removed) and 0 otherwise.  
We then perform a channel-wise max-pooling across the global encoding $e_g$ and the attention-weighted local encoding $P_\ell$, as $P = \max (e_g, P_\ell)$.

\vspace{-0.5em}
\subsubsection{Multi-Level Decoder}
\vspace{-0.5em}
Our decoder consists of multiple decoder streams that work in parallel to decode the fused embedding $P$ (Figure~\ref{fig:model_diagram}). The global decoder $D_{g}$ takes the embedding $P$ as input and attempts to generate an entire completed point cloud $Y_g$.
In parallel, we use a local decoder $D_{\ell}$ to decode the points in each region of the input space. The embedding $P$ is concatenated with a one-hot vector indicating each region's location and create a region-specific embedding. Through one-hot encoding, each decoder specializes in completing a certain region and learns a region-specific embedding. The decoder takes as input these region-specific embeddings and generates a subset of the output point cloud localized to the respective region $Y_\ell^i$. The generated local regions are combined together to obtain the full point cloud $Y_\ell$.
The multi-level output generated by the network captures the details of the object at global and local levels. The outputs of the multi-level decoder streams, $D_{\ell}$ and  $D_{g}$, are concatenated to form the final prediction of our network as $Y$.



\vspace{-1em}
\subsection{Point Cloud Completion Losses}
\label{sec:losses}
\vspace{-0.5em}
The standard loss used for comparing two point clouds is the Chamfer Distance (CD). It is a bi-directional permutation invariant loss over two point clouds representing the nearest neighbor distance between each point and its closest point in the other cloud. In our method, 
we use an asymmetric Weighted Chamfer Distance loss, $\mathcal{L}_{wcd}$, defined as, 
\begin{equation}
\footnotesize
\mathcal{L}_{wcd} (X, Y) = \frac{(1 - \beta)}{|X|}\sum_{x \in X} \min_{y \in Y}\|x - y\|_2  +  
    \frac{\beta}{|Y|}\sum_{y \in Y} \min_{x \in X} \|y - x\|_2
\label{equ:cd_global_loss}
\end{equation}

\noindent where $X$ is the original partial point cloud used here as pseudo-ground truth and $Y$ is the output. 
Importantly, we only compute the loss for the regions that are present in $X$. A weight of $(1-\beta)$ is applied to the first term in Eqn.~\ref{equ:cd_global_loss} which penalizes the distance from each point in $X$ to its nearest neighbor in $Y$. This term enforces that the output point cloud $Y$ should contain points that are close to those in $X$. Note that the input to the network is $\hat{X}$, which has synthetic occlusions, not $X$, which is the original partial point cloud. A weight of $\beta$ is applied to the second term in Eqn.~\ref{equ:cd_global_loss} to penalize the distance from each point in $Y$ to its nearest neighbor in $X$. 
We do not expect this term to reach 0 for a well-trained network since $X$ only contains a partial point cloud, while output $Y$ contains the entire point cloud; we still find it a helpful regularization. 
We impose the following variants of $\mathcal{L}_{wcd}$ on the model,
\textbf{Inpainting-Global Loss:} This loss acts as a \textit{global shape loss}, focusing on the overall shape of an object. We impose it as the Weighted Chamfer Distance (Eqn.~\ref{equ:cd_global_loss}) between original partial point cloud $X$ and output of the global decoder $Y_{g}$ and define it as $\mathcal{L}_{wcd} (X, Y_{g})$. 
\textbf{Inpainting-Local Loss:} We impose Inpainting-Local loss as the Weighted Chamfer distance between each region output $Y_\ell^i$ from local decoder $D_{\ell}$ and corresponding partitioned region $X_i$ in the original partial point cloud $X$ where $i$ indexes over regions. While Inpainting-Global loss considers the entire $X$ to find the nearest neighbor, Inpainting-Local loss differs in that it only considers the partitioned region $X_i$ to find the nearest neighbor. Thus, it acts as a \textit{local shape loss} that enables the network to learn region-specific shapes and embeddings and focus on the finer details of an object. We do not penalize regions that are missing in $X$ where a region is considered missing if the number of points in that region is below a certain threshold. The Inpainting-Local Loss is therefore defined as, $\sum_{i}\mathbbm{1}_i \cdot \mathcal{L}_{wcd} (X_{i}, Y_\ell^i)$ where the indicator function $\mathbbm{1}_i$ equals one if region $i$ is present in $X$ and zero otherwise. 
\textbf{Multi-View Consistency:} Similar to Gu~\etal~\cite{gu2020weakly}, our method uses multi-view consistency as an auxiliary loss. Their method explicitly performs pose estimation. Similarly, we perform an estimated pose canonicalization (weakly supervised, Sec.~\ref{sec:experimental_setup}). We also show later~(Sec.~\ref{sec:robustness}) that our method is robust to canonicalization errors. During training, we sample a view $k$ from $V$ partial views of an object $X$ given as $X^{1}, \ldots, X^{V}$. Since all the views of an object correspond to the same object, for input partial point cloud $X^{k}$, the loss is computed with all views $X^{1}, \ldots, X^{V}$. We define a global inpainting multi-view consistency loss as $\sum_{j=1}^{V}\mathcal{L}_{wcd} (X^{j}, Y_{g}^{k})$ where $X^{j}$ is the $j^{th}$ view of $X$ and $Y_{g}^{k}$ is the global output from decoder $D_{g}$. We also define local inpainting multi-view consistency loss as $\sum_{i}\sum_{j=1}^{V}\mathbbm{1}^j_i \cdot \mathcal{L}_{wcd} (X^{j}_{i}, Y_\ell^{i,k})$ where $i$ indexes over regions, $X^{j}_{i}$ is the $i^{th}$ region of view $X^j$, 
$Y_\ell^{i,k}$ is the region output from local decoder $D_{\ell}$ for input $X^{k}_{i}$, and $\mathbbm{1}^j_i$ is $1$ if region $i$ is present in $X^{j}$ and $0$ otherwise.
During training, we sum the losses as, $\sum_{j=1}^{V}\mathcal{L}_{wcd} (X^{j}, Y_{g}^{k}) + \sum_{i}\sum_{j=1}^{V}\mathbbm{1}_i \cdot \mathcal{L}_{wcd} (X^{j}_{i}, Y_\ell^{i,k})$. This multi-view information is only available at training time.

\vspace{-1.5em}
\begin{figure*}[t]
 	\centering
 	\includegraphics[width=0.7\textwidth]{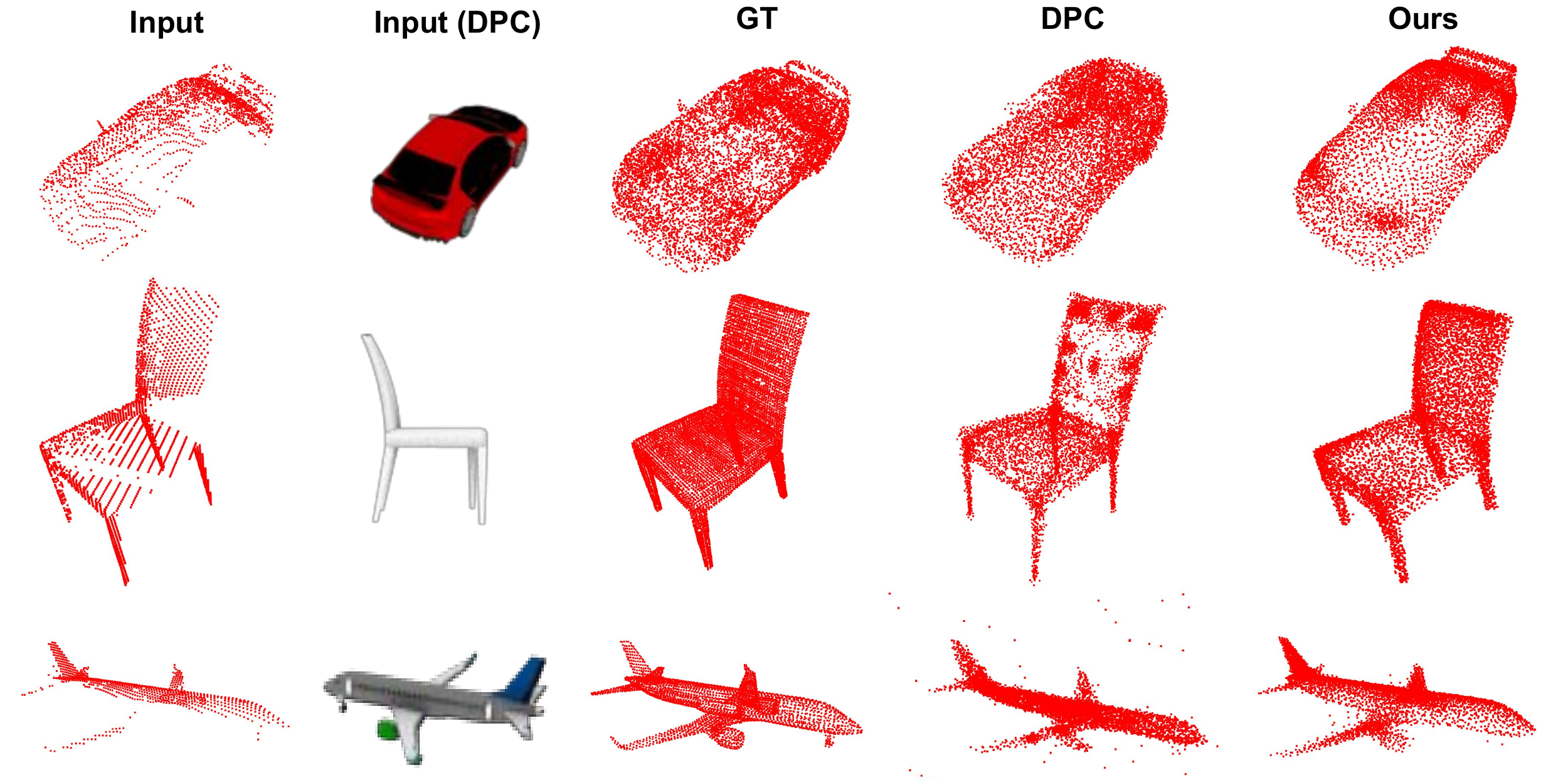}
 	\vspace{-1em}
 	\caption{Qualitative results on the ShapeNet dataset compared with our baseline, DPC~\cite{insafutdinov2018unsupervised}. Our method is better able to reconstruct fine-grained object details~(back portion of the car and engines on the airplane), produces fewer noisy points for the airplane and produces more uniformly distributed points in the chairs than the baseline.} 
 	\label{fig:qualtitative_results}
 	\vspace{-2em}
\end{figure*} 
\section{Experiments}
\vspace{-0.5em}

\label{sec:experiments}
\subsection{Implementation Details}
\vspace{-0.5em}
During test-time, we use a single view of an object. The multiple views are only available during training. To get the final completed point cloud, we concatenate the output from multi-level decoders $D_g$ and $D_{\ell}$. We do not remove regions at inference time. Otherwise, the network during inference is the same as described above.
For consistency with prior work~\cite{gu2020weakly}, we resample each partial point cloud $X$ to have a total of 3096 points.
PointPnCNet uses architecture from PCN~\cite{yuan2018pcn} for encoder and decoder blocks. The model is trained from scratch for 400K iterations with batch size of 32, learning rate of 5e-4 decayed by 0.5 after every 100K iterations and $\beta = 0.25$ in $\mathcal{L}_{wcd}$. Please refer to appendix for more details.

\vspace{-1em}
\subsection{Experimental Setup}
\vspace{-0.5em}
\label{sec:experimental_setup}
Following the evaluation protocol of Gu~\etal~\cite{gu2020weakly}, we test our approach on ShapeNet~\cite{chang2015shapenet} and Semantic KITTI~\cite{behley2019semantickitti}. ShapeNet has ground truth annotations for each object class which allows us to evaluate how well our method generates completed shape. On the other hand, Semantic KITTI allows us to evaluate the robustness of our method on real LiDAR data. 

The observations are transformed to a canonical frame (a shared reference frame which aligns all instances of a class) using canonical frame predictions generated via IT-Net~\cite{yuan2018iterative} for ShapeNet and predicted bounding boxes obtained from OpenPCDet~\cite{openpcdet2020} for Semantic KITTI. We use IT-Net for ShapeNet as it is trained in a weakly-supervised manner from only classification labels and learns to align the instances of each class, without any pose supervision. In general, any pose estimator can be used here. We evaluate the robustness of our method to this canonical frame estimation in Sec.~\ref{sec:robustness}.



\textbf{ShapeNet:} ShapeNet~\cite{chang2015shapenet} is a synthetic dataset with 3D CAD models. We report our results on three categories, airplanes, cars, and chairs, that are commonly used in the related works~\cite{insafutdinov2018unsupervised, gu2020weakly}.  We use the same data split provided by DPC~\cite{insafutdinov2018unsupervised},  where RGB-D data is generated for random camera views with fixed translation, similar to Gu \etal~\cite{gu2020weakly}. For evaluation, we use ground truth point clouds provided by DPC~\cite{insafutdinov2018unsupervised} which are densely sampled from ShapeNet meshes and downsampled to 8192 points.

\textbf{Semantic KITTI:} We evaluate our method for a real-world scenario using KITTI~\cite{behley2019semantickitti}. Previous methods~\cite{xie2020grnet, gu2020weakly, wen2020point, yuan2018pcn} have a standard protocol of evaluation on real-world data by testing on the cars of KITTI only. We adopt the same protocol in our work. Following Gu \etal~\cite{gu2020weakly}, we train over the parked car instances~(which have multiple views captured when a LiDAR sensor moves through the scene and scans a parked car from different locations) with sequences 00 to 10 (excluding 08) as train set and sequence 08 as test set. The train set consists of 507 parked car instances and 46152 observations, while the test set has 229 parked car instances and 16296 observations.

Although having no complete ground truth information in KITTI creates some limitations in its evaluation, testing on this dataset shows the ability of our method to handle real-world LiDAR data. By combining the evaluations on a real-world dataset~(KITTI) and a synthetic dataset~(ShapeNet), which has ground truth annotations, we are able to present a more thorough evaluation. This is the standard evaluation procedure following Gu \etal~\cite{gu2020weakly}.



\textbf{Metrics:} 
Our primary metric for quantitatively evaluating shape completion is the \textit{Chamfer Distance (CD)}, as is used in previous works~\cite{gu2020weakly,yuan2018pcn,wang2020cascaded}. We define this metric in its weighted form in Equation \ref{equ:cd_global_loss}. For evaluation, to compare with the ground-truth completed point cloud, we equally weight each component with a $\beta$ of $0.5$.  
Additionally, we follow Gu \etal~\cite{gu2020weakly} and report each component of the Chamfer distance independently: the mean distance from each predicted point to its nearest true point described as \textit{Precision}, and the mean distance from each true point to its nearest predicted point described as \textit{Coverage}. \textit{Precision} describes how well the predicted points match the local shape, while \textit{Coverage} is related to how much of the shape is completed. We also evaluate the Earth Mover's Distance~(EMD)~\cite{yuan2018pcn}, which finds a bijection between the predicted point cloud and the ground truth point cloud that minimizes the average distance between corresponding points. Like previous work, we also evaluate the F-score@1\%~\cite{xie2020grnet}.






\vspace{-1em}
\subsection{Point Cloud Completion Results}
\vspace{-0.5em}
We compare with the current state-of-the-art unsupervised methods, DPC~\cite{insafutdinov2018unsupervised} and Gu~\etal~\cite{gu2020weakly}. 
Table~\ref{table:shapenet_main_results} shows our method outperforming the baseline methods on the synthetic ShapeNet dataset, producing lower Chamfer distances across all shape categories. \textit{Precision} and \textit{Coverage} metrics also improve, showing that our method produces more accurate points and better covers the full object shape.
Our method is also able to outperform DPC~\cite{insafutdinov2018unsupervised} as per the Earth Mover Distance (EMD) metric (Table~\ref{table:kitti_agg_emd}b).
Since the code for \cite{gu2020weakly} is not publicly available, the EMD metric on that method cannot be evaluated.

We further show in Table~\ref{table:agg_ablation_sem_kitti}a that our method outperforms the previous state-of-the-art~\cite{gu2020weakly} on the Semantic KITTI dataset, generating outputs that are significantly more accurate than~\cite{gu2020weakly}.
The KITTI dataset is more realistic than ShapeNet. With samples having a range of sparsity (since real-world LIDAR gets sparse with distance), it represents the data available in self-driving scenarios. We also show improvement compared to a simple densification baseline (Densified Input in Table~\ref{table:agg_ablation_sem_kitti}a) which suggests that our method is indeed completing the partial point clouds rather than simply densifying them. This densification baseline uniformly samples points within the volume of a local surface that is approximated as an ellipsoid, formed using eigenvalues for 10 nearest neighbors of each point in input partial point cloud. We also conduct a uniformity analysis whose results we report in the appendix.


\begin{table*}[t]
\centering
\footnotesize
\setlength{\tabcolsep}{3pt}
\resizebox{0.85\textwidth}{!}{
\begin{tabular}{c|ccc|ccc|ccc}
\multirow{2}{*}{\textbf{Method}}            & \multicolumn{3}{c|}{\textbf{Airplane}}                                                                              & \multicolumn{3}{c|}{\textbf{Car}}                                                                                   & \multicolumn{3}{c}{\textbf{Chair}}                                                                              \\ \cline{2-10} 
                                                     & \textbf{CD}                       & \multicolumn{1}{c}{\textbf{Precision}} & \multicolumn{1}{c|}{\textbf{Coverage}} & \textbf{CD}                       & \multicolumn{1}{c}{\textbf{Precision}} & \multicolumn{1}{c|}{\textbf{Coverage}} & \textbf{CD}                       & \multicolumn{1}{c}{\textbf{Precision}} & \multicolumn{1}{c}{\textbf{Coverage}} \\ \hline \hline
 \multicolumn{1}{l|}{DPC \cite{insafutdinov2018unsupervised}}     & \multicolumn{1}{c}{3.91}                              & \multicolumn{1}{c}{-}               & \multicolumn{1}{c|}{-}              & \multicolumn{1}{c}{3.47}                              & \multicolumn{1}{c}{-}               & \multicolumn{1}{c|}{-}              & \multicolumn{1}{c}{4.30}                             & \multicolumn{1}{c}{-}              & \multicolumn{1}{c}{-}           \\                                         
\multicolumn{1}{l|}{Gu~\etal~\cite{gu2020weakly}}     & \multicolumn{1}{c}{1.95}                              & \multicolumn{1}{c}{0.91}               & \multicolumn{1}{c|}{1.05}              & 2.68                              & \multicolumn{1}{c}{1.27}               & \multicolumn{1}{c|}{\textbf{1.41}}              & 3.33                              & \multicolumn{1}{c}{1.69}               & \multicolumn{1}{c}{1.64}           \\ \hline
 \multicolumn{1}{l|}{PointPnCNet~(Ours)}                                & \multicolumn{1}{c}{\textbf{1.66}}          &    \multicolumn{1}{c}{\textbf{0.79}}                                  &         \multicolumn{1}{c|}{\textbf{0.87}}                              & \multicolumn{1}{l}{\textbf{2.48}}          &                 \multicolumn{1}{c}{\textbf{0.99}}                       &    \multicolumn{1}{c|}{1.49}                                    & \multicolumn{1}{c}{\textbf{2.70}} &           \multicolumn{1}{c}{\textbf{1.36}}                             &            \multicolumn{1}{c}{\textbf{1.34}}                        \\ \hline
\end{tabular}
}
\caption{Quantitative results on the Airplane, Cars, and Chairs categories of the ShapeNet dataset. All results are multiplied by a factor of 100, following Gu~\etal~\cite{gu2020weakly}.}
\label{table:shapenet_main_results}
\vspace{-1em}
\end{table*}

\begin{table}[t]
\small
\centering
\setlength{\tabcolsep}{2pt}
\begin{tabular}{ cc } 
\resizebox{0.4\textwidth}{!}{
\begin{tabular} {l|ccc}
\textbf{Method} & \textbf{CD} & \textbf{Precision} & \textbf{Coverage} \\
\hline
\hline
Gu~\etal~\cite{gu2020weakly} & 0.194 & 0.087 & 0.107 \\
Densified Input & 0.130  & \textbf{0.025}  & 0.105   \\
\hline
PointPnCNet (Ours) & \textbf{0.095}  &  0.045 &  \textbf{0.050} \\ \hline
\end{tabular}} &
\resizebox{0.57\textwidth}{!}{
\begin{tabular}{l|cccc} 
 & \multicolumn{4}{c}{\textbf{Model without}} \\ \cline{2-5}
 \textbf{Dataset} &  \textbf{Inpainting} & \textbf{Multi-View Loss} & \textbf{Global Loss} & \textbf{Local Loss} \\
\hline\hline
ShapeNet & +0.98 & +0.27 & +0.50 & +0.32 \\
KITTI & +0.05 & +0.03 & +0.30 & +0.24 \\ \hline

\end{tabular}}  \\
\footnotesize{(a)} & \footnotesize{(b)} \\
\end{tabular}
\caption{(a). Quantitative results on the Semantic KITTI dataset, (b). Increase in mean Chamfer distance on ShapeNet and KITTI datasets for various ablations of our method. ShapeNet results are averaged across each object category.}
\vspace{-1em}
\label{table:agg_ablation_sem_kitti}
\end{table}

\begin{table}[!h]
\small
\centering
\setlength{\tabcolsep}{2pt}
\begin{tabular}{ cc} 
\multirow{4}{*}[1em]{
      \includegraphics[width=0.4\textwidth]{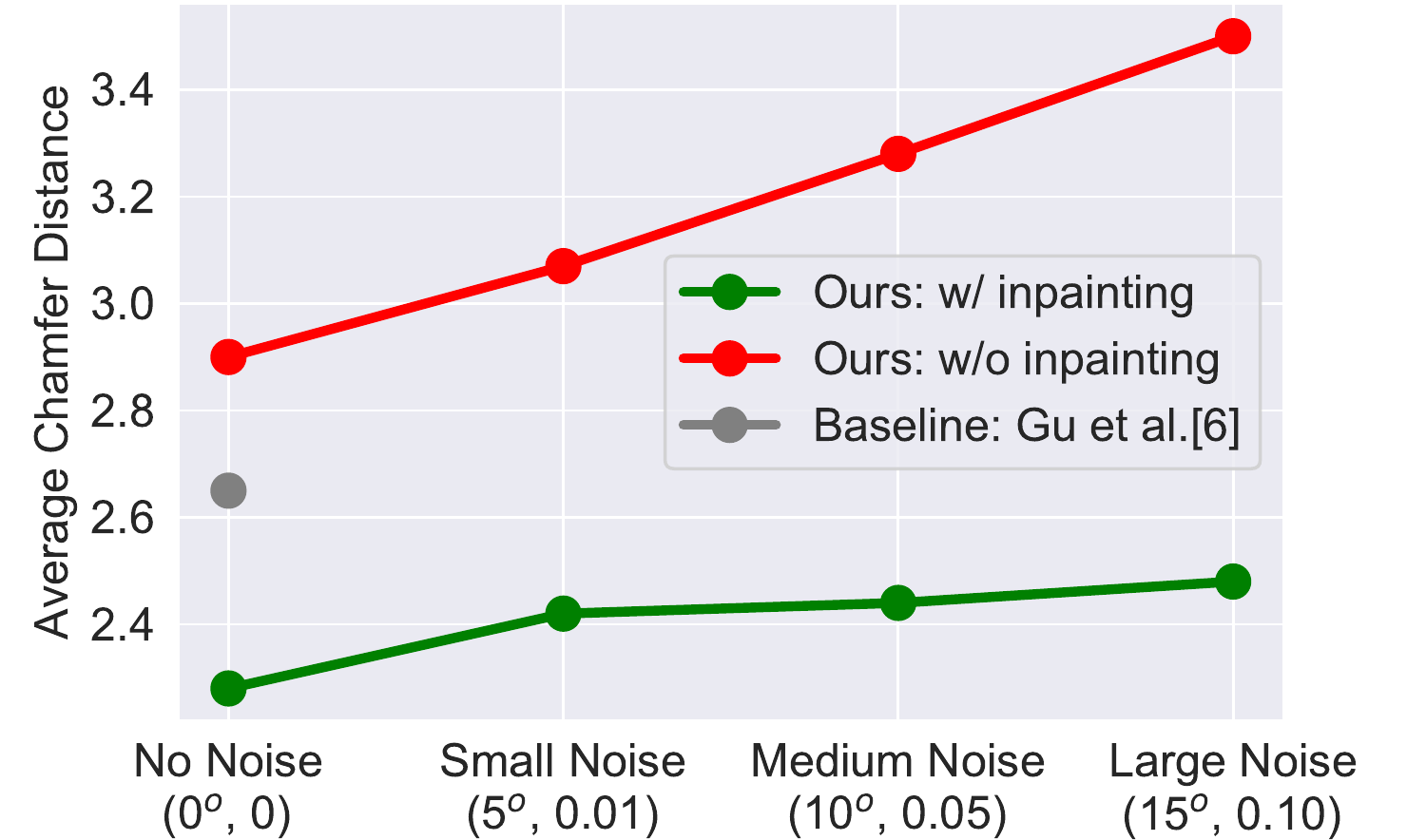}
} &
\resizebox{0.4\textwidth}{!}{
\begin{tabular} {l|ccc}
\textbf{Training Method}    & \textbf{Airplane} & \textbf{Car} & \textbf{Chair} \\ \hline \hline
\multicolumn{1}{l|}{DPC \cite{insafutdinov2018unsupervised}} & 0.146 & 0.160 & 0.194 \\
\multicolumn{1}{l|}{PointPnCNet (Ours)} & \textbf{0.108}  &  \textbf{0.120} &  \textbf{0.140} \\ \hline
\end{tabular}
} \\
& \footnotesize{(b)} \\ \\
& \resizebox{0.4\textwidth}{!}{
\begin{tabular}{c|ll}
\textbf{Training Method} & \multicolumn{1}{c}{\textbf{F-score@1\%} $\uparrow$ } & \multicolumn{1}{c}{\textbf{EMD} $\downarrow$} \\ \hline \hline
 \multicolumn{1}{l|}{PointPnCNet (Ours)}       &    \textbf{0.67}                                      &                         \textbf{0.45}      \\ 
 \multicolumn{1}{l|}{Ours~w/o inpainting}           &            0.44                              &                 0.62                 \\
\hline  
\end{tabular}
}  \\ 
\footnotesize{(a)} & \footnotesize{(c)} \\
\end{tabular}
\caption{(a). Mean of the chamfer distance across ShapeNet categories (Airplane, Car, Chair). Our method is trained with noisy poses, with \& without inpainting, shown in \textcolor{green}{green} and \textcolor{red}{red}, respectively. Baseline Gu~\etal~\cite{gu2020weakly} has no added noise, (b). Earth Mover Distance (EMD) metric on Shapenet. Lower EMD is better, (c). F-Score@1\% and EMD metrics on Semantic KITTI. We evaluate them on our method vs our ablation of without inpainting.}
\vspace{-1.625em}
\label{table:kitti_agg_emd}
\end{table}

\vspace{-1.125em}
\subsection{Qualitative Results}
\vspace{-0.5em}

\begin{figure*}[t]
 	\centering
 	\includegraphics[width=0.8\textwidth]{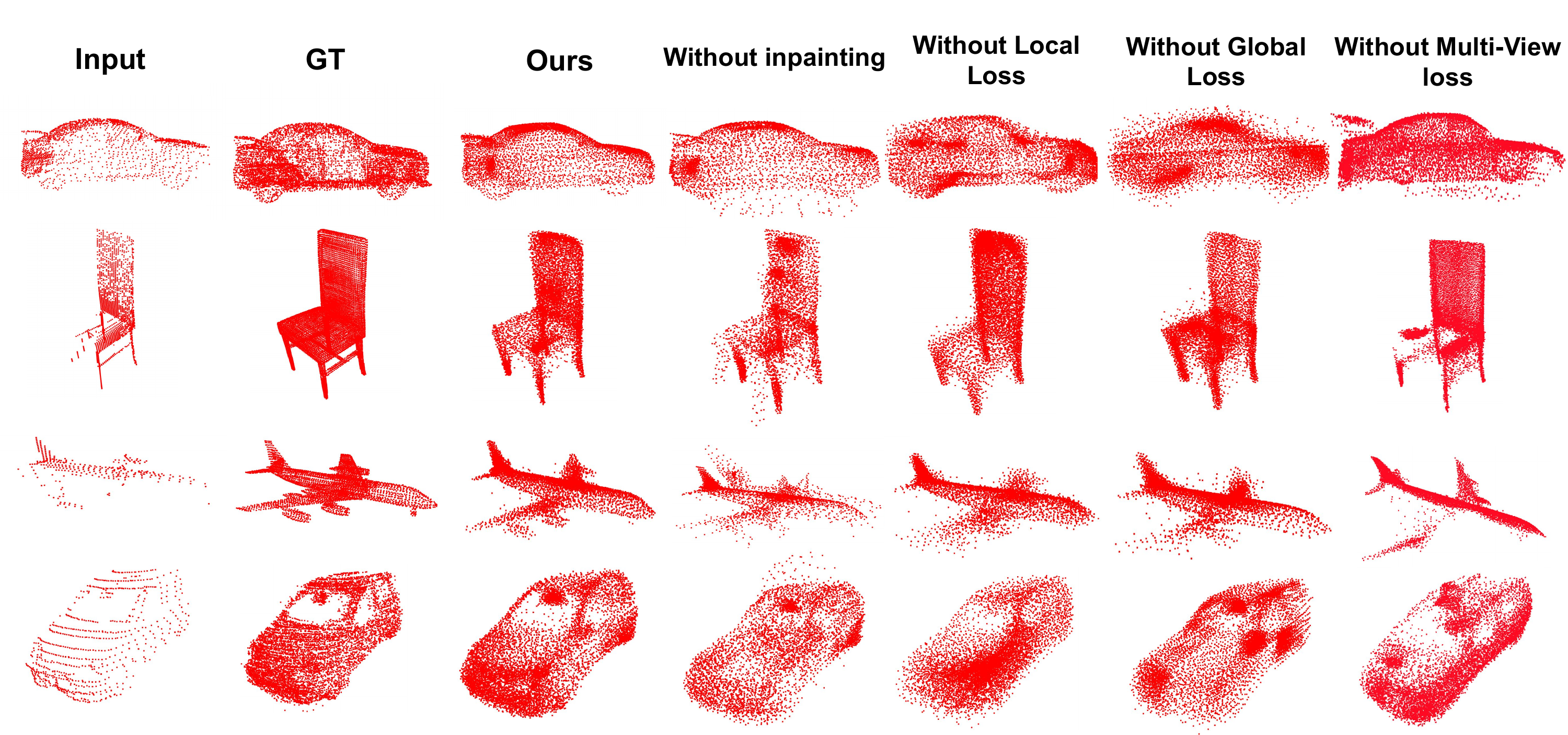}
 	\vspace{-1em}
 	\caption{Qualitative results for ablation study on ShapeNet and KITTI. Without inpainting, local loss, global loss, and multi-view loss, the network yields noisy output.}
 	\label{fig:ablation_results}
  	\vspace{-18pt}
\end{figure*}

We present the qualitative results of our method for each category of ShapeNet in Figure~\ref{fig:qualtitative_results} and KITTI in Figure~\ref{fig:ablation_results}. In comparison to the baseline DPC~\cite{insafutdinov2018unsupervised}, we observe that our method is able to better cover the target object with a more uniform distribution over the target surface and accurately reconstructs the fine-grained object details. For example, our method is able to complete the back of the car and the side mirror whereas the baseline outputs noisy points. For chairs, our method generates more uniformly distributed points whereas the baseline outputs patches/clusters of points in that location. This highlights the fact that our method is better able to generalize and complete the unseen regions of incomplete shapes.

\vspace{-1.125em}
\subsection{Ablation Study}
\vspace{-0.5em}

\noindent \textbf{Inpainting Loss:} Region removal to create synthetic occlusions and the task of inpainting are removed, \textbf{Multi-View Loss:} The multi-view loss is removed from our training method. Each partial point cloud is used to supervise its own, synthetically occluded completion, \textbf{Global Level:} We remove the Inpainting-Global Loss, global encoder $\mathbf{E_g}$ and global decoder $\mathbf{D_g}$ from our completion pipeline. \textbf{Local Level:} The Inpainting-Local loss, local encoder $\mathbf{E_{\ell}}$ and local decoder $\mathbf{D_{\ell}}$ are removed from our method. The number of output points for the \textit{global ghape} and \textit{local shape} ablations are kept consistent with our full method.

We report the ablation results in Table~\ref{table:agg_ablation_sem_kitti}b on ShapeNet, as an average over all categories, and on Semantic KITTI. We find that all components of our system are crucial for optimal performance across both datasets. We further report the F-score@1\%~\cite{xie2020grnet} and the EMD metric on the Semantic KITTI dataset with the ablation of removing inpainting in Table \ref{table:kitti_agg_emd}c. We find that inpainting greatly improves our results across both of these metrics.



The qualitative effects of our ablation study can be seen in Figure~\ref{fig:ablation_results}. We observe that inpainting generates an object-specific, less noisy output, when comparing "Ours" and "Without Inpainting". Our method without local loss fails to complete local details of an object such as back of a car or wings of a plane and without the global loss predicts a generic, noisy shape of an object. Since each local encoder and decoder only observe the points within that region and not the points in the other potentially occluded regions, they allow the network to focus on individual parts of an object and be robust to different occlusion patterns. The local loss helps in creating a more uniform completion, since it completes its associated region and the global loss reasons about the entire shape of the object. Finally, without multi-view loss, the output point cloud is noisy and incomplete as can be seen in all shapes.
\vspace{-1.125em}
\subsection{Robustness to Canonical Frame Estimation}
\vspace{-0.5em}
\label{sec:robustness}
Previous work~\cite{gu2020weakly} depends on multiple views which can be sensitive to the pose alignment errors. While we also use a multi-view loss, our inpainting losses make the model robust to noisy alignment allowing it to learn from poorly aligned data. The mean rotation/translation difference, after using IT-Net for pose canonicalization, between the multiple partially observed shapes during both training and inference is $5.46^{\circ}/~0.008$, $12.33^{\circ}/~0.013$ and $7.12^{\circ}/~0.010$ for Car, Chair, and Plane, respectively~(unit of translation is object diameter). This shows that even the canonicalized poses are not perfectly aligned and due to inpainting, our method is still able to learn from this poorly aligned data (particularly with respect to rotation). To further highlight the contribution of inpainting to this robustness, we add noise to the predicted IT-Net poses with rotations and translations sampled uniformly with a maximum displacement of 5\textdegree / 0.01, 10\textdegree / 0.05, and 15\textdegree / 0.10. Figure in Table~\ref{table:kitti_agg_emd}a shows that without inpainting (in \textcolor{red}{red}), our method is extremely sensitive to alignment noise but with inpainting (in \textcolor{green}{green}), our method only degrades slightly with higher noise, and remains more accurate at all levels of noise than the baseline~\cite{gu2020weakly} with no noise added.

\vspace{-1.125em}
\subsection{Impact of $\beta$ in the Weighted Chamfer Distance loss} 
\vspace{-0.5em}

\begin{wraptable}[9]{r}{6cm}
\vspace{-14pt}
\tiny
\centering
\setlength{\tabcolsep}{2pt}
\begin{tabular}{ c}
\resizebox{0.4\textwidth}{!}{
\begin{tabular}{c|llll}
\textbf{$\beta$} & \multicolumn{1}{c}{\textbf{Airplane}} & \multicolumn{1}{c}{\textbf{Car}} & \multicolumn{1}{c}{\textbf{Chair}} & \multicolumn{1}{c}{\textbf{KITTI}} \\ \hline \hline
0    &      2.10                                 &      2.63                            &   3.02                                 &   0.132                              \\
0.25 &      \textbf{1.66}                                 &              \textbf{2.48 }                   &       \textbf{2.70}                             &          \textbf{0.095 }                         \\
0.5  &      2.31                                 &      3.00                            &       3.28                             &      0.121                              \\
0.75 &          2.59                             &              3.50                    &               3.78                     &              0.142                      \\
1    &      3.83                                 &  4.72                                &    4.95                                &         0.196                           \\ \hline
\end{tabular}} \\ 
\end{tabular}
\caption{Performance analysis on different values of hyperparameter $\beta$ used in Equation~\ref{equ:cd_global_loss}.}
\label{table:beta_table}
\end{wraptable}
We present an analysis in Table~\ref{table:beta_table} in which we train PointPnCNet with different values of $\beta$ to understand the contribution of the second term in the asymmetric Weighted Chamfer Distance loss (Eqn.~\ref{equ:cd_global_loss}), $\mathcal{L}_{wcd}$.
 


From Table~\ref{table:beta_table}, we can observe that 
the optimal performance occurs at $\beta = 0.25$ across both the ShapeNet and KITTI datasets.
Our intuition is that a larger value of $\beta$ imposes a penalty for generating points in $Y$ in the regions that were occluded in the input; this contradicts our goal of completing those missing regions.  Nonetheless, setting $\beta = 0$ also leads to worse performance because the second term in Eqn.~\ref{equ:cd_global_loss} is needed to (minimally) penalize the network for predicting points in $Y$ that are far from the original partial point cloud $X$. Setting $\beta = 0.25$ provides the appropriate balance between these competing objectives.  As explained in Section~\ref{sec:losses}, this tradeoff only occurs for the global loss; the local loss uses a regional indicator that only applies the loss to regions for which we have ground truth information.

\vspace{-1.5em}
\section{Conclusion}
\vspace{-0.5em}

We propose a self-supervised method for point cloud completion via inpainting and random region removal that can be trained using only LiDAR-based partial point clouds. Our method produces significantly more accurate point cloud completions and outperforms the previous unsupervised methods on ShapeNet and Semantic KITTI. Through exhaustive ablation, we show the importance of each component of our method and the robustness to alignment errors. While the current method uses intersecting half-spaces defined by coordinate planes, other methods for point cloud partitioning can be explored in future work. We hope that our method will improve real-world 3D object understanding.
\vspace{-1em}
\section{Acknowledgements}
\vspace{-0.5em}

This material is based upon work supported by the National Science Foundation under Grant No. IIS-1849154, and the CMU Argo AI Center for Autonomous Vehicle Research.

\bibliography{egbib}

\section{Appendix}

\subsection{Architecture Details}
For all results and ablations, we keep the output size of our network as 8192 points, where the global decoder $D_{g}$ generates 4096 points and the local decoder $D_{\ell}$ generates 512 points for each region, to make an overall size of 4096 points across all local regions. Similarly, the input size is kept consistent for all the ablations; that is, the input size is 3096 and 387 for global encoder $E_{g}$ and local encoder $E_{\ell}$ respectively. Each region in $X$ is dropped with a probability of removal of 20\% and the resulting synthetically occluded point cloud $\hat{X}$ is passed to the global encoder $E_{g}$.  In parallel, the input partial point cloud is subdivided into 8 regions along the axial planes of the canonical frame. Each region not artificially removed or marked as missing is then independently encoded using the local encoder, $E_{\ell}$. When encoding each region of the input cloud, regions that are marked as missing based on the threshold number of points are replaced with zeros equal to the threshold. In our method, we set this threshold as 4. We allow a small overlap of 0.02 cm between neighboring regions for the ShapeNet dataset and 0.02m for the KITTI dataset. The architecture of local encoder $E_{g}$ and global decoder $D_{g}$ are similar to the PCN~\cite{yuan2018pcn}.  For local encoder $E_{\ell}$ and local decoder $D_{\ell}$, we use the architecture of PCN~\cite{yuan2018pcn}, but reduce the number of hidden units to 1/8th of the original number. We use Adam optimizer with a learning rate of $1 \times 10^{-4}$ and train our network for 400K iterations.

\subsection{Data preparation}

\textbf{Shapenet}: We obtain a point cloud from the RGB-D data by backprojecting 2.5D depth images to 3D similar to Gu \etal~\cite{gu2020weakly}. In contrast to DPC~\cite{insafutdinov2018unsupervised}, we do not use the color information. The centers of the oriented clouds are then shifted to the origin before passing it to our shape completion network. Specifically, we use the 3D partial shape classification branch of IT-net pre-trained on ModelNet40 to generate the pose transformations, as it does not require the ground-truth pose annotations for training. Since our method does not require perfect pose alignment, using IT-Net pretrained on ModelNet40 instead of ShapeNet is sufficient for our purpose, as it represents an off-the-shelf canonical frame estimator for our model classes.  We refer the reader to IT-Net~\cite{yuan2018iterative} for details on this pose canonicalization method.

Originally, the ShapeNet~\cite{chang2015shapenet} dataset has 5 views. When training on $N$ views, we only consider a fixed set of $N$ random views, which is chosen at the beginning of training; the network is only trained on these $N$ views and the other views of an object are discarded.

\textbf{Semantic KITTI:} At training time, we subdivide the observations of a single instance into groups of 20 sequential observations and randomly sample a set of four views for multi-view training. When evaluating accuracy on this dataset, all 20 frames are combined using ground truth odometry to form the ground truth shape of each instance. This merged cloud is only used for evaluation and is not present during training. At inference time, only a single view is used.

\subsection{Ablation Studies}
In this section, we present a more exhaustive ablation study focusing on the number of views, architecture changes, number of input points used for training and mention the details of the ablation of densification of input point clouds for the KITTI dataset.

\subsubsection{Number of views}
We evaluate the sensitivity of our method to the number of views available at training time in Supplementary Figure~\ref{fig:views_plot}.  We show the results both with and without inpainting in \textcolor{green}{green} and \textcolor{red}{red} lines respectively. It can be observed that our model is able to outperform the baseline with 2 views and 3 views, even though the baseline Gu~\etal~\cite{gu2020weakly} is trained with 4 views.  This demonstrates that our method is able to take advantage of a reduced number of views, due to our use of inpainting. 
We also show the qualitative results with varying numbers of training views in the Supplementary Figure~\ref{fig:views_results}; as can be seen, the results of 2 views and 3 views are qualitatively very similar to the results with 4 views. 


\subsubsection{Architecture Changes}

\paragraph*{\textbf{Global and Local Encoders and Decoders}}
We analyze whether to use both global and local encoders and decoders in our network. The results can be found in Supplementary Table~\ref{tbl:shapenet_ablation}. It can be observed that a combination of global and local encoders and decoders gives the best performance among all the possible combinations.


\paragraph*{Number of levels} In addition to the two levels in our parallel model (global and local), we experiment with adding another branch where the partial point cloud is partitioned into $3\times 3\times 3$ regions. For this branch, we use an independent local encoder and decoder. The input size of a region to the encoder is taken as 115 points (to maintain a total input size of 3096) and the size of the predicted point cloud is 152 points for each region (to maintain a total output size of 4096 for the local decoder).  For computing the loss, we divide the original input (before dropping points) into regions and subsample the points to have at most 304 points in each region. The results are in Supplementary Table~\ref{tbl:more_ablations}. We notice that further partitioning of the partial point cloud and the additional branch do not give a significant improvement in the performance.


\subsubsection{Number of input points}
\begin{figure}[hbt!]
\centering
 \includegraphics[trim=0 0 0 0, clip,width=0.6\textwidth]{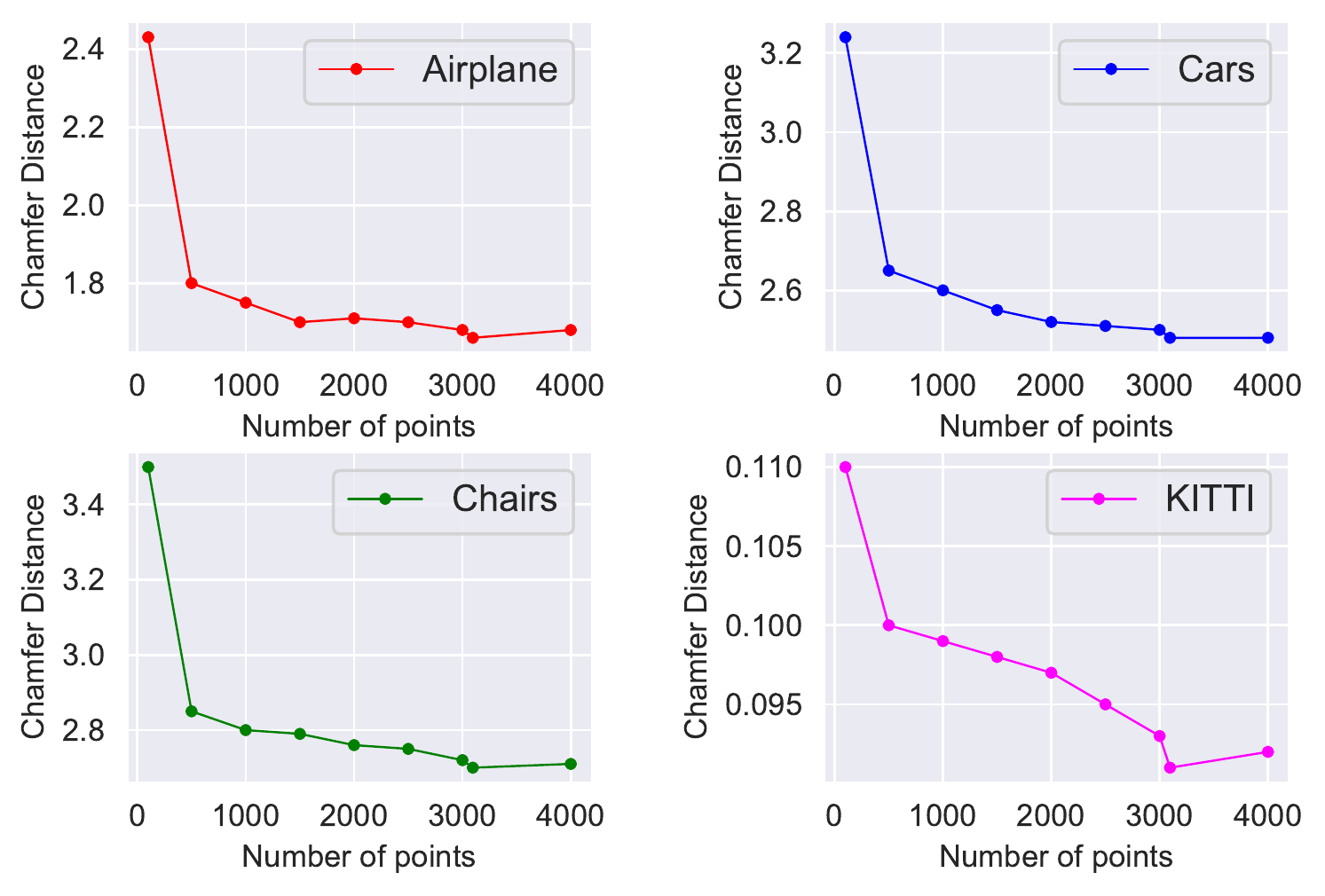}
 \caption{Quantitative Results of the Chamfer Distance metric with respect to the number of points in the input point cloud during testing.}
 \label{fig:numpoints_chamfer}
\end{figure}

We evaluate the effect of the number of points in the point cloud on the performance of our method.  To test this, we create new versions of the test set with varying numbers of points; for each object, we resample the point cloud (without replacement) from the input point cloud with a varying number of sampled points.
We evaluate the Chamfer Distance metric as a function of the number of points in the input point cloud on the ShapeNet and KITTI~\cite{behley2019semantickitti} dataset during testing. We evaluate our method on the number of points ranging from 100 to 4000 and present the results in Figure~\ref{fig:numpoints_chamfer}. As expected, performance degrades as we reduce the number of available points.

\subsubsection{Densification of KITTI point clouds}
\label{sec:Densification of KITTI point clouds}
To evaluate the quantitative effects of simply densifying the input point cloud without completing occluded regions, we design a simple densification method. For each point in the input partial point cloud, we find its 10 nearest neighbors and estimate the eigenvalues of this local neighborhood. An ellipsoid is formed using these values and points are uniformly sampled within this volume. This approximates the local surface. From Table 2 of the main paper, the improvement of our method over the results of this densification method demonstrates that our model is completing the partial point clouds rather than simply densifying the partial input cloud.

\subsubsection{Performance Analysis with respect to Occlusions}
We conduct an experiment to assess the impact of occlusions in the input partial point clouds on the ability of the model to complete the given shape. To do so, we introduce artificial occlusions by removing a certain number of regions from the input during testing (we have divided the input into 8 total regions). Given that the original input is already naturally occluded, we artificially remove at most three regions because beyond that, the input is barely visible. The results are shown in Table~\ref{tbl:num_region}; we can observe that as the number of artificial occlusions in the input increases, there is a slight drop in performance for all categories. However, the model is considerably robust to the additional occlusions.

\subsection{Metrics}
In this section, we report different metrics for further analysis of our method.

\subsubsection{Precision and Coverage of observed and unobserved regions}
For a detailed analysis, we compute the precision and coverage of the observed and unobserved regions of the input point cloud. To categorize points as observed or unobserved, we compute the distance between each point in the predicted point cloud and its nearest neighbor in the input point cloud. We compute the mean and standard deviation of these distances for each point cloud and use 1 standard deviation over this mean as a threshold. Points with the nearest neighbor distance greater than this threshold are considered as unobserved, while all other points are considered observed. The precision and coverage are computed separately for each of these types of points and we report the results in the Supplementary Table~\ref{tbl:obs_unobs}. As expected, we find that the precision and coverage of the observed regions are slightly better than that of unobserved regions in the input partial point cloud; however, the results are relatively similar for the observed and unobserved regions, which provides further evidence that we are completing (and not just densifying) the input (see also Section~\ref{sec:Densification of KITTI point clouds}).

\subsubsection{F1-Score}
Following Xie et al~\cite{xie2020grnet}, we evaluate the F1-score@1\%, which is the harmonic mean between precision and recall, on the ShapeNet dataset. In this context, ``precision" is the percentage of the points in the predicted point cloud which are within a specified distance threshold with the ground truth. ``Recall" is the percentage of the points in the ground truth point cloud that are within a distance threshold with the predicted point cloud. Precision helps to measure the accuracy of the prediction and recall measures the coverage of the prediction. In this metric, we use ${d = 1\%}$ of the side length of the predicted point cloud. It can be observed from the Supplementary Table~\ref{tbl:shapenet_f1score} that our method is able to outperform the baseline DPC~\cite{insafutdinov2018unsupervised} when evaluated on this metric. We do not report the results on Gu~\etal~\cite{gu2020weakly} since their code is not open-source.

\begin{figure}[t]
\centering
 \includegraphics[width=0.5\textwidth]{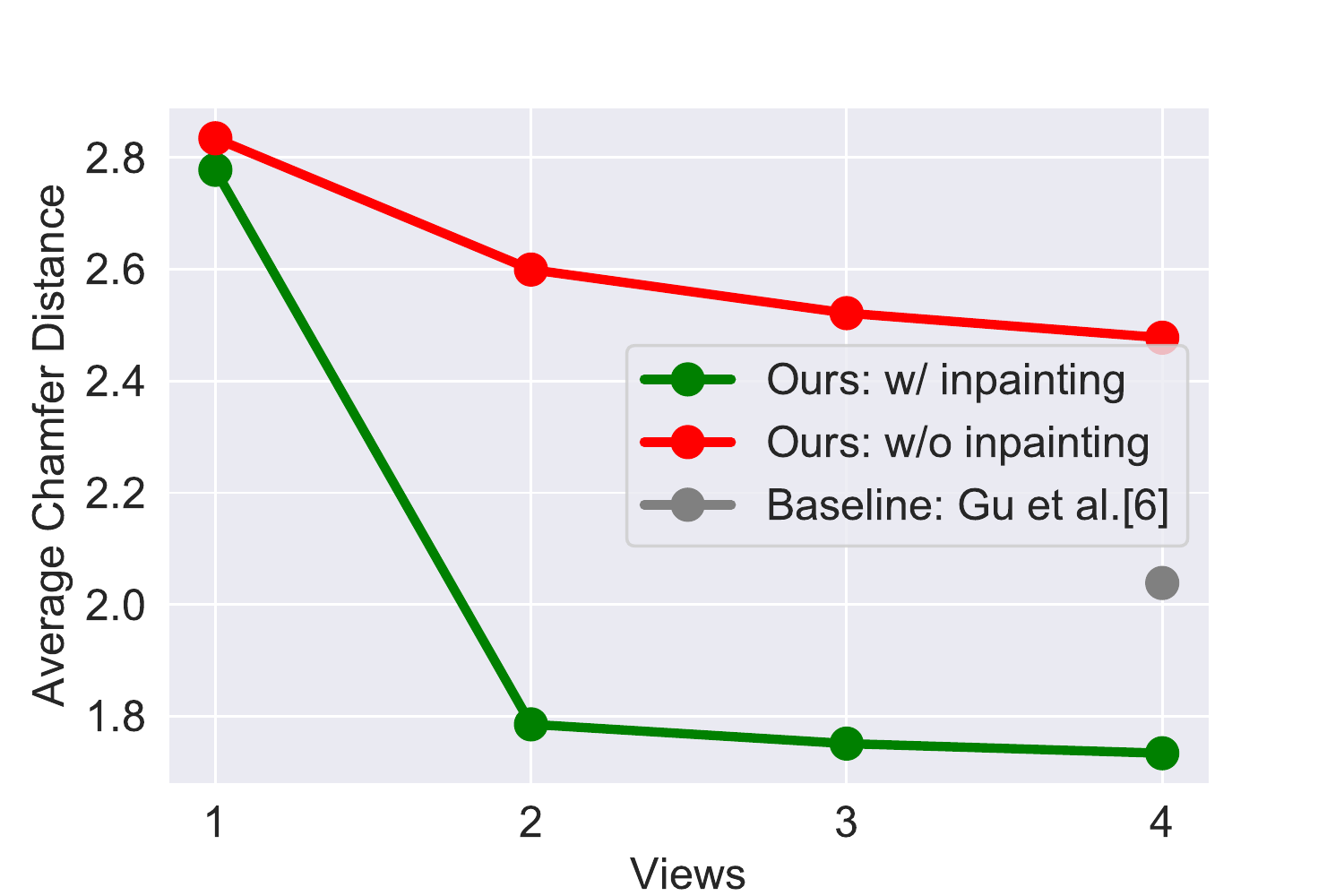}
\caption{Quantitative Results on the number of views (1, 2, and 3) (with \textcolor{green}{green} and without inpainting \textcolor{red}{red}) used during network training. Our original method trains on 4 views. All the values reported are average Chamfer Distance metric over the ShapeNet~(Airplane, Car, Chair) and KITTI dataset. We are able to outperform the baseline using a limited number of views due to our use of inpainting.}
 \label{fig:views_plot}
\end{figure} 

\begin{figure*}[t]
\centering
\includegraphics[width=\columnwidth]{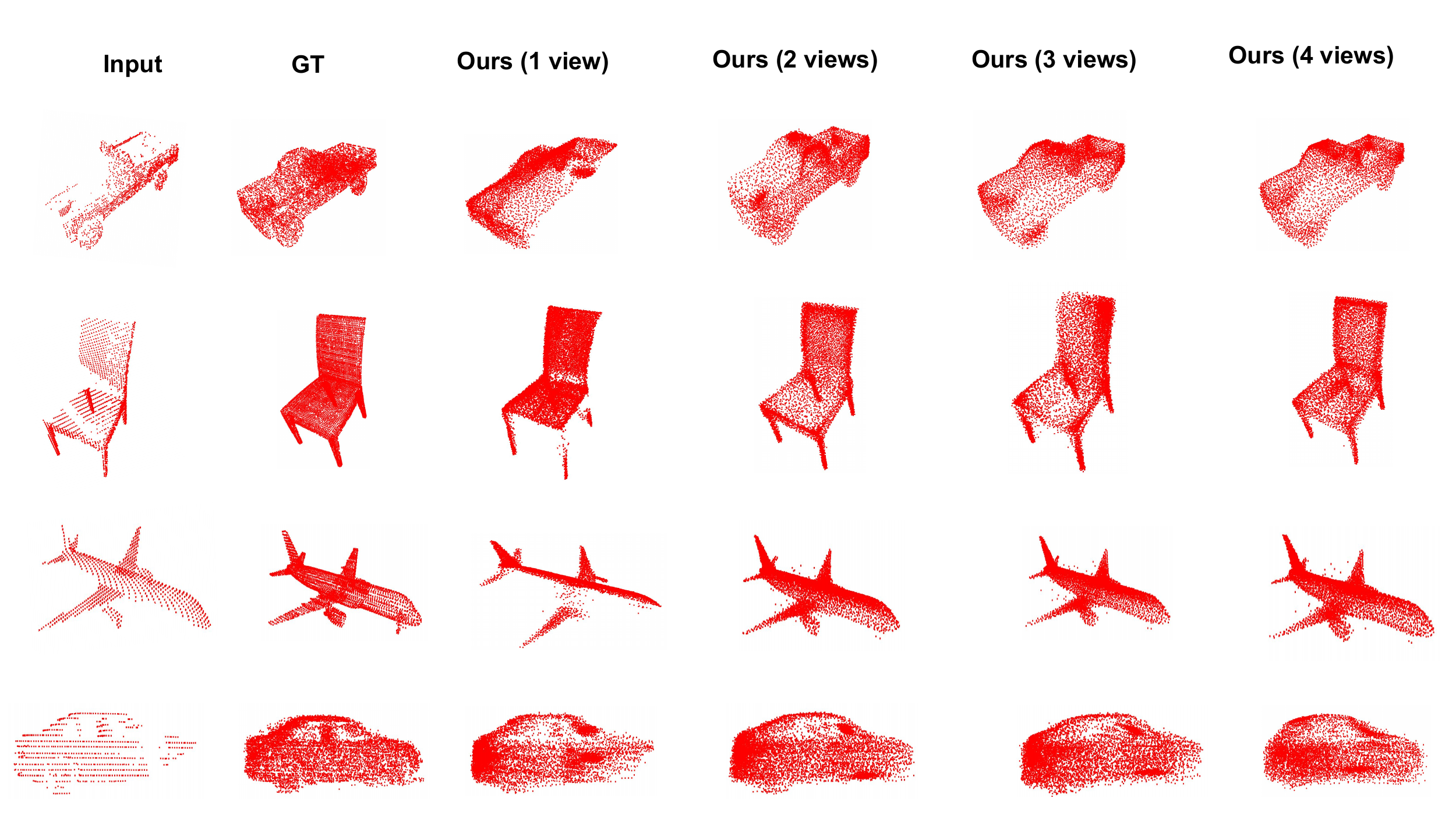}
 \caption{Qualitative results on varying the number of views given as input to the PointPnCNet. The first, second, third, and fourth row shows the results on the ShapeNet test set of car, chair, plane, and Semantic KITTI~\cite{behley2019semantickitti} dataset respectively. As can be seen, the results of 2 views and 3 views are qualitatively very similar to the results with 4 views. This demonstrates that our method is able to take advantage of a reduced number of views, due to our use of inpainting.}
 \label{fig:views_results}
\end{figure*} 

\subsubsection{Uniformity Metric}
We also evaluate the uniformity metric following Xie et al~\cite{xie2020grnet} on the ShapeNet and KITTI datasets. In the Supplementary Table~\ref{tbl:shapenet_uniformity}, we compare our method with the baseline DPC on the ShapeNet dataset. Our method gives a similar performance with the baseline with respect to this metric, revealing that both methods have similar uniformity of predicted points. 

For the KITTI dataset, we compare our method with the ablation of our method without inpainting, as DPC does not train and evaluate on KITTI and Gu~\etal~\cite{gu2020weakly} do not have open-source code. We report the results on KITTI in Supplementary Table~\ref{tbl:kitti_uniformity} and show the improvement in the performance of our model when using inpainting.



\subsection{Qualitative Results}

We present additional visualizations of the complete predicted point cloud generated by our network, PointPnCNet.

\begin{figure*}[hbt!]
 	\centering
 	\includegraphics[width=0.99\textwidth]{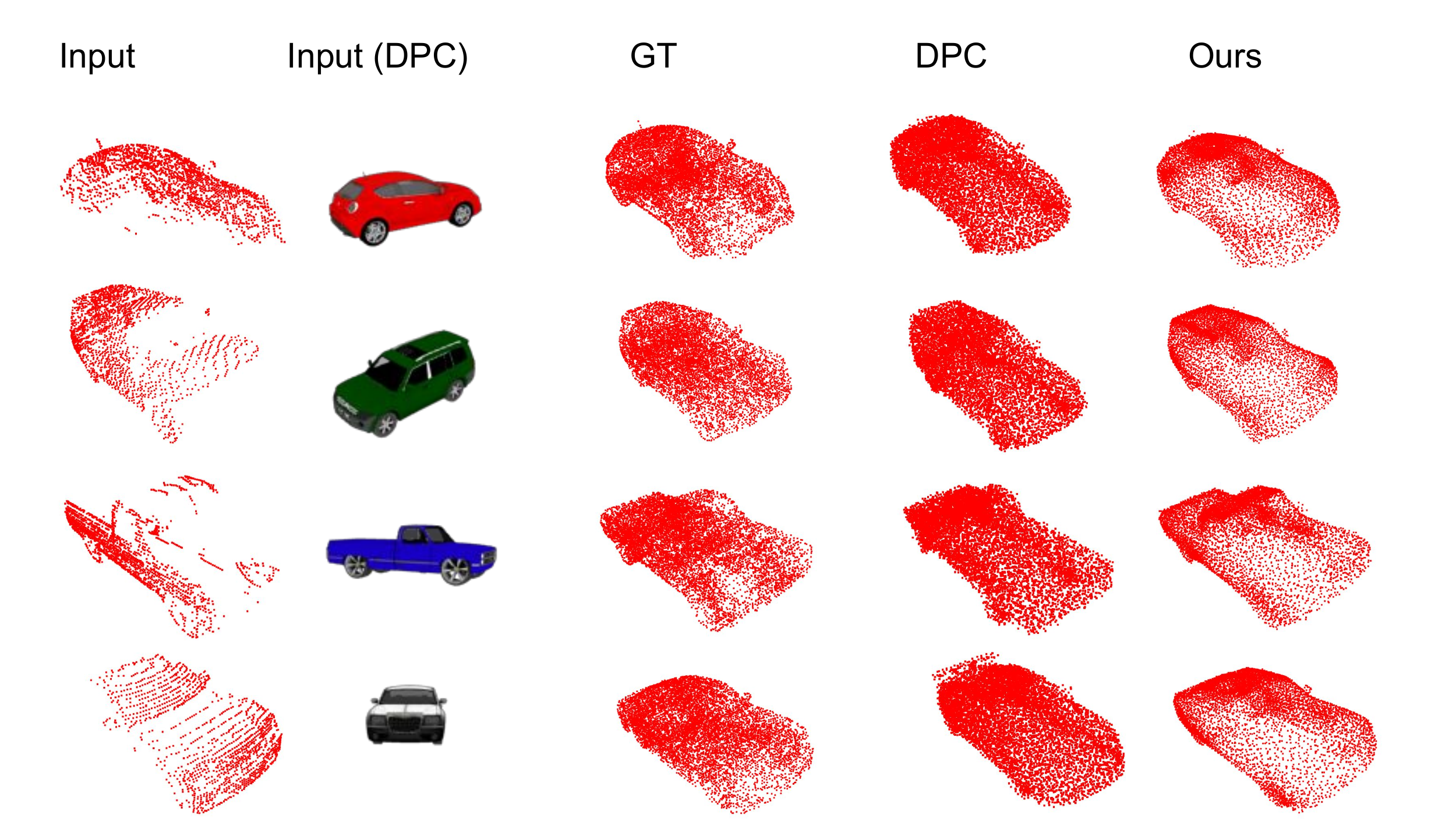}
 	\caption{Success cases on the Car category of the ShapeNet dataset. It can be observed that our model is able to complete the finer details of a car such as the headlight of a car and generates a detailed outer boundary in comparison to DPC~\cite{insafutdinov2018unsupervised} in all the rows. It is also able to generate the shape of a truck as can be seen in the third row.}       
 	\label{fig:shapenet_car_success}
\end{figure*}

\textbf{Cars:} As can be observed from the Supplementary Figure~\ref{fig:shapenet_car_success}, our model is able to complete the finer details of a car such as the headlight of a car and generates a more defined outer boundary in comparison to DPC~\cite{insafutdinov2018unsupervised}. We also show that our network has the ability to not only complete the shapes of general cars, but also the shape of a truck as shown in the third row of Supplementary Figure~\ref{fig:shapenet_car_success}. We show a few failure cases as well on the car category in the Supplementary Figure~\ref{fig:shapenet_car_fail}. Our method is unable to create detailed shapes of various sports cars. Further, for the truck in the second row, our method fails to create a gap between the front and back of a truck.

\textbf{Chairs:} We present in Supplementary Figure~\ref{fig:shapenet_chair_success} that our method is able to generate finer completion results on different types of chairs such as a sofa and desk chair than DPC~\cite{insafutdinov2018unsupervised}. It is able to complete the front, back, and arms of the chair. There are also a few failure cases where the network generates noisy results especially near the legs of a chair as seen in Supplementary Figure~\ref{fig:shapenet_chair_fail}. 

\begin{figure*}[hbt!]
 	\centering
 	\includegraphics[width=0.99\textwidth]{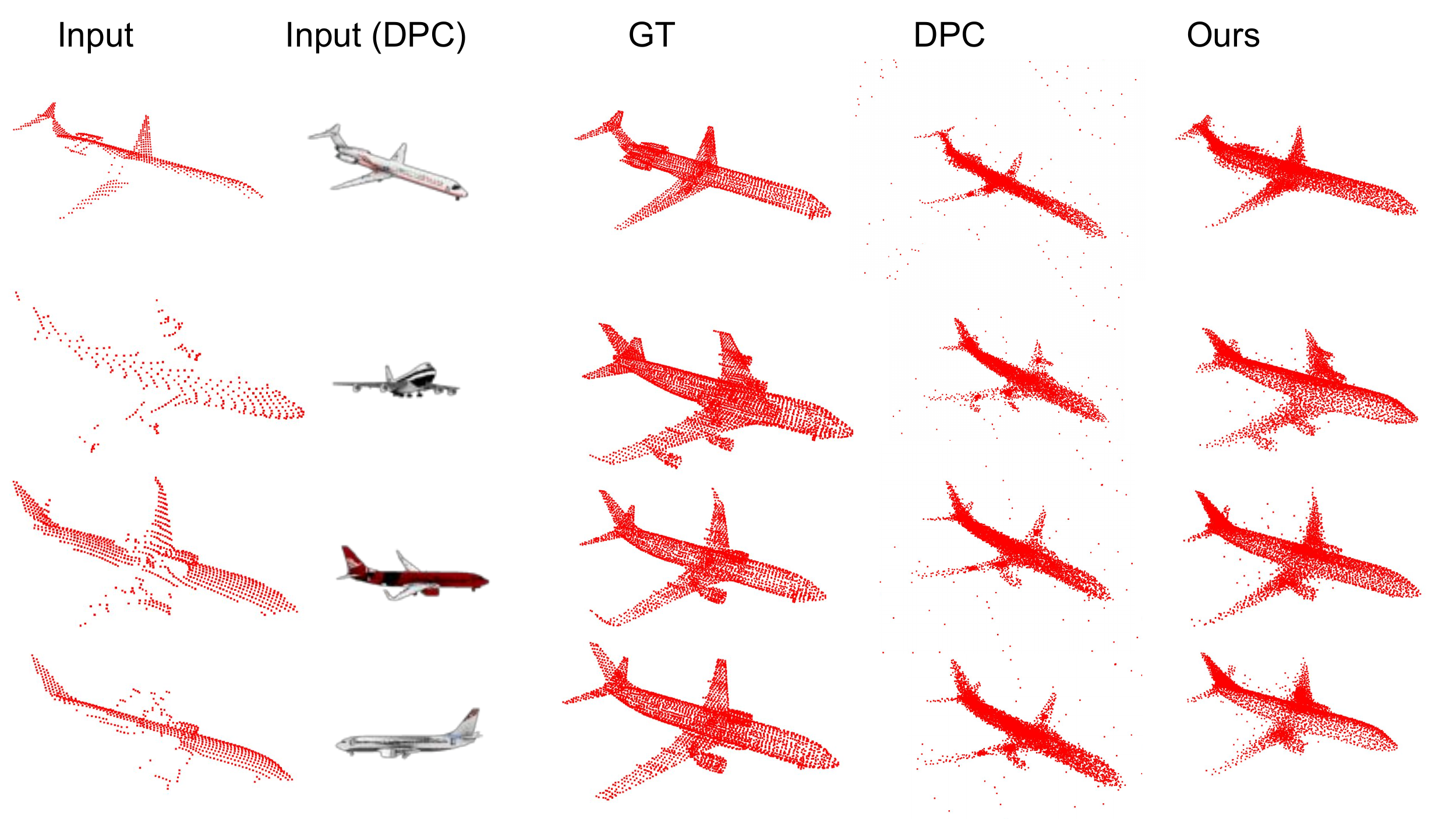}
 	\caption{Success cases on the Plane category of the ShapeNet dataset. We compare our method with the results of DPC~\cite{insafutdinov2018unsupervised}. It can be observed that the network is able to complete the front, back and wings of the planes.}       
 	\label{fig:shapenet_plane_success}
\end{figure*}

\begin{figure*}[hbt!]
 	\centering
 	\includegraphics[width=0.99\textwidth]{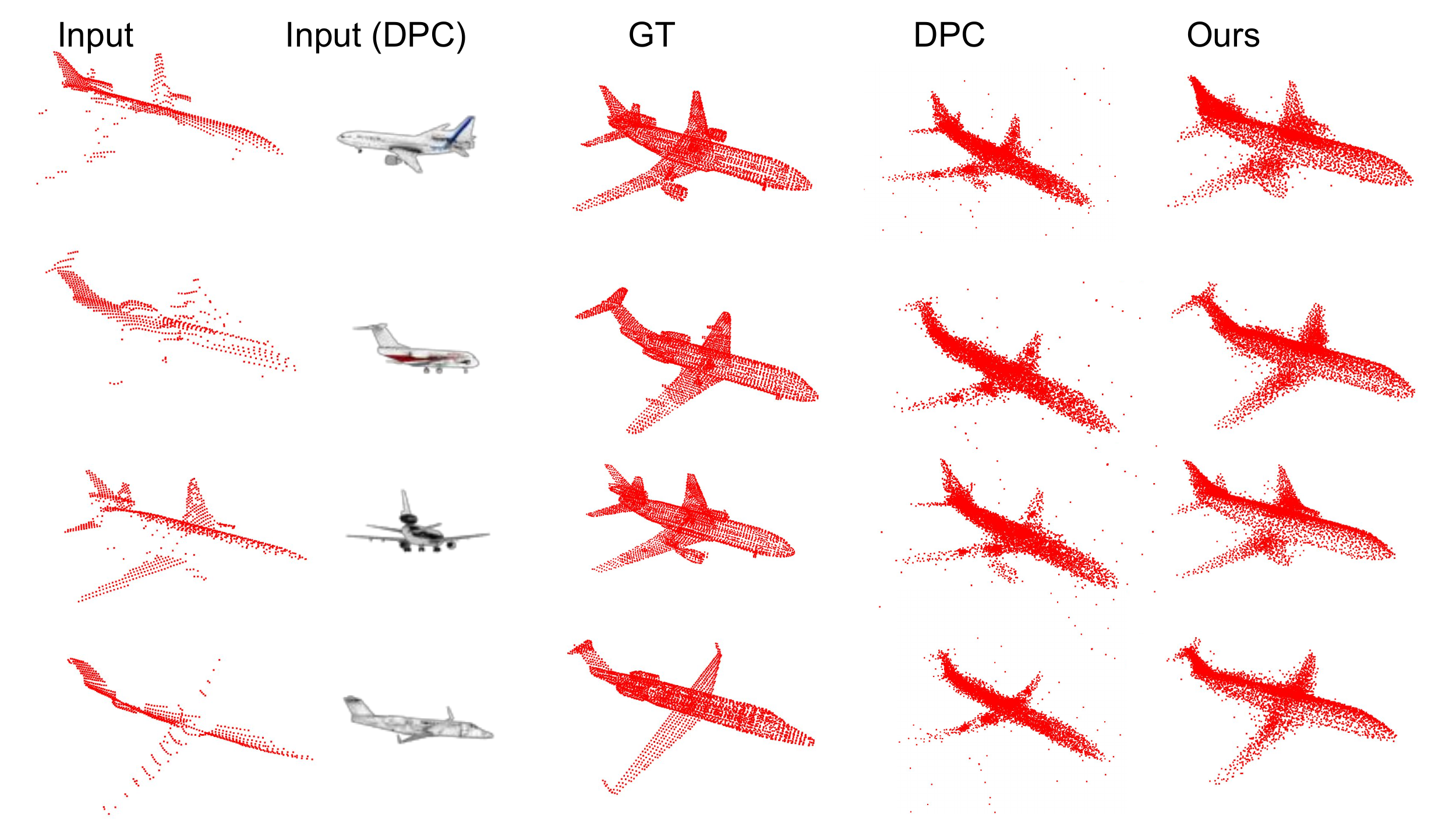}
 	\caption{Failure cases on the Plane category of the ShapeNet dataset. We compare our method with the results of DPC~\cite{insafutdinov2018unsupervised}. The network generates some noisy points near the wings of the planes.}       
 	\label{fig:shapenet_plane_fail}
\end{figure*}

\textbf{Airplanes:} 
From Supplementary Figure~\ref{fig:shapenet_plane_success}, we observe that the network is able to complete the front, back, and wings of the planes. Supplementary Figure~\ref{fig:shapenet_plane_fail} shows some failure cases in which it also generates some noisy points near the wings of the planes.

\textbf{KITTI:}  We show the visualizations where our network is able to complete the partial point cloud cars from the LiDAR scans of the Semantic KITTI dataset in the first and second row of Supplementary Figure~\ref{fig:kitti_supp}. Additionally, there are a few failure cases where the network is unable to generate the details in a fine manner such as the tire of a car as seen in the third and fourth row of Supplementary Figure~\ref{fig:kitti_supp}. We also show the completion results of the partial point clouds in a scene in the Supplementary Figure~\ref{fig:kitti_scene}.

\begin{figure*}[hbt!]
 	\centering
 	\includegraphics[width=0.99\textwidth]{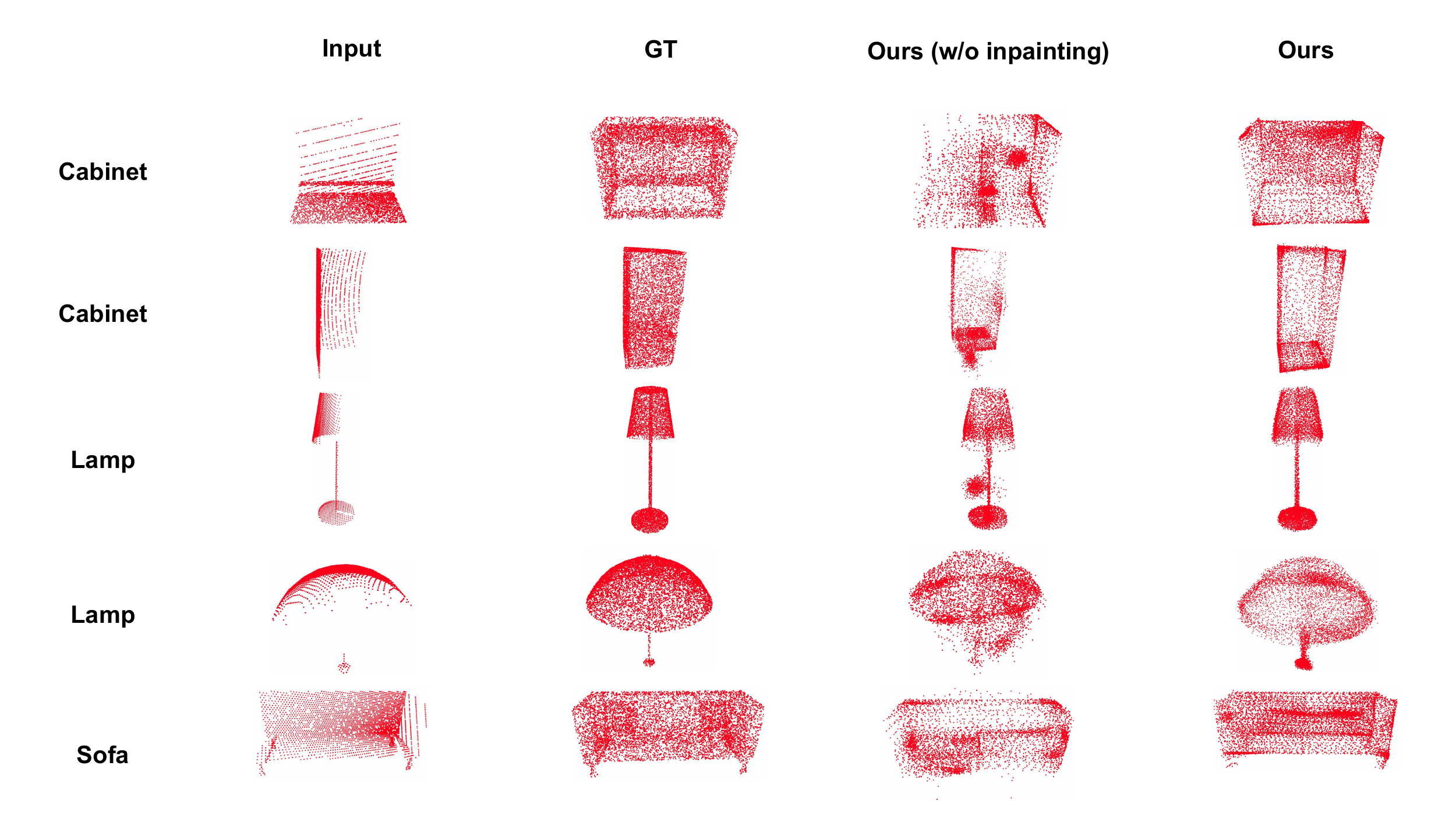}
 	\includegraphics[trim=0 0 0 50, clip,width=0.99\textwidth]{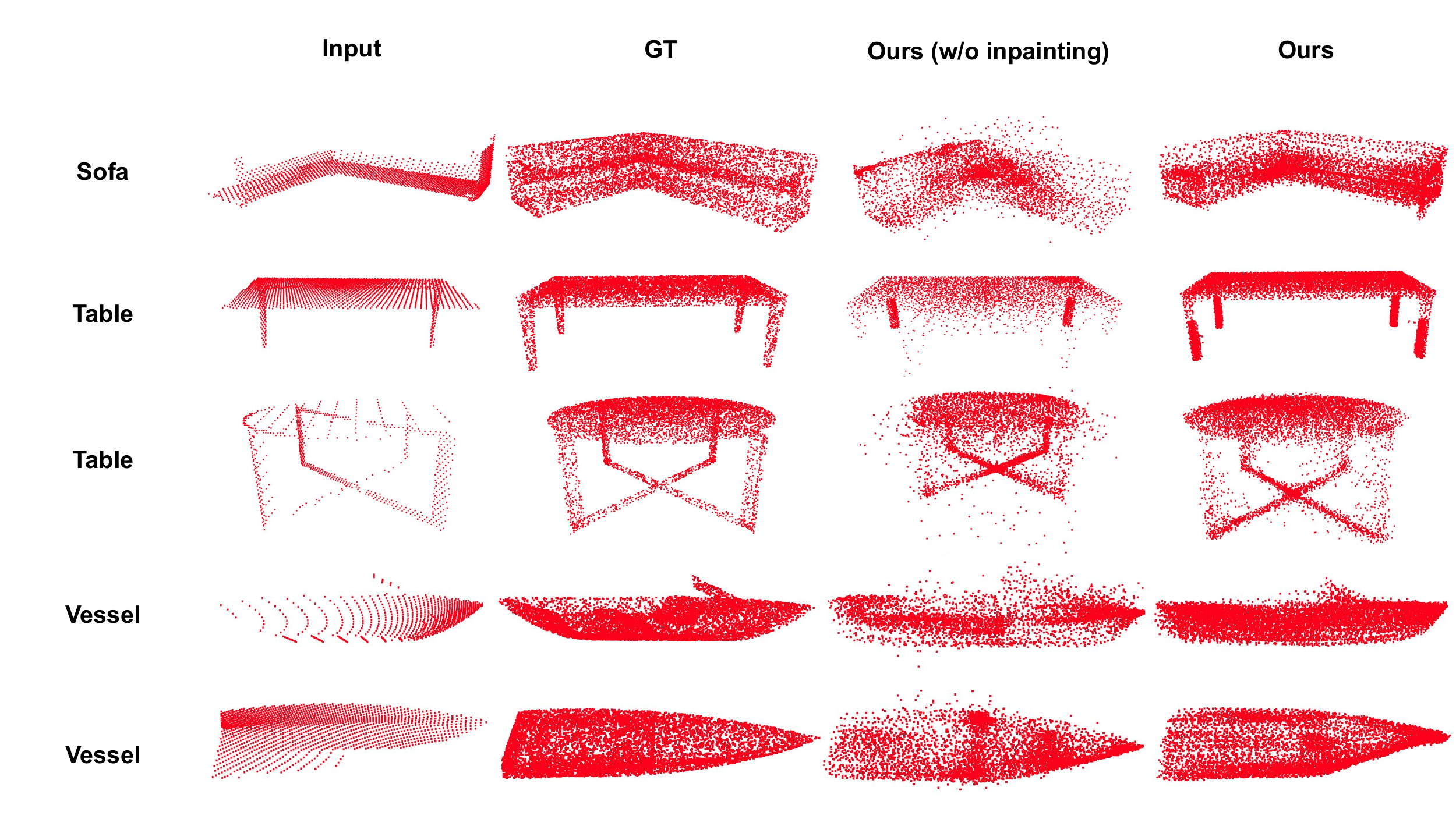}
 	\caption{Qualitative Results on five categories of Shapenet compared to our ablation of without inpainting.}       
 	\label{fig:shapenet_categories}
\end{figure*}

\textbf{ShapeNet Categories: } We present the qualitative results on the 5 other categories of the ShapeNet dataset - Cabinet, Lamp, Sofa, Table, and Vessel in the Figure~\ref{fig:shapenet_categories}. We compare the results of our method with our ablation of without inpainting. It can be observed that our method is able to complete the shape of the incomplete point clouds whereas our method without inpainting outputs noisy points.

\subsection{Comparison with supervised method}
To analyze the performance gap between self-supervised method and supervised method, we compare the performance of our method with a fully supervised method, PCN~\cite{yuan2018pcn} on 8 categories of the ShapeNet dataset and present the results in Table~\ref{tbl:pcn_sup_self_sup}. Since our method builds on the architecture of PCN, we compare our method to fully-supervised PCN; the choice of architecture is somewhat orthogonal to our proposed method of inpainting. We observe that the fully supervised PCN outperforms our self-supervised method, as expected. However, our results indicate that our method has reduced the gap between self-supervised and fully supervised approaches. In Table~\ref{tbl:pcn_sup_self_sup}, we also compare our method to the ablation of ``no inpainting" across 8 object categories of ShapeNet and show consistent improvement in performance.

\begin{table}[hbt!]
\centering
{
\begin{tabular}{cccc||c|c|c|c}
\hline
\multirow{2}{*}{$E_{g}$} & \multirow{2}{*}{$E_{\ell}$} & \multirow{2}{*}{$D_{g}$} & \multirow{2}{*}{$D_{\ell}$} & \textbf{Airplane}         & \textbf{Car}          & \textbf{Chair}        & \textbf{KITTI}       \\ \cline{5-8} 
                              &                               &                               &                               & \textbf{CD}               & \textbf{CD}           & \textbf{CD}           & \textbf{CD}          \\ \hline
\checkmark &  &  \checkmark  &              & \multicolumn{1}{c|}{1.830} & \multicolumn{1}{c|}{2.710} &   \multicolumn{1}{c|}{3.260} & \multicolumn{1}{c}{0.336}  \\ 
\checkmark &  &              & \checkmark   & \multicolumn{1}{c|}{1.930} & \multicolumn{1}{c|}{2.560} &  \multicolumn{1}{c|}{3.480}  & \multicolumn{1}{c}{1.042} \\ 
\checkmark &  &  \checkmark  & \checkmark   & \multicolumn{1}{c|}{1.820} & \multicolumn{1}{c|}{2.580} & \multicolumn{1}{c|}{3.320} & \multicolumn{1}{c}{0.357} \\ 
\hline
& \checkmark  &  \checkmark  &              &    \multicolumn{1}{c|}{1.950}   &   \multicolumn{1}{c|}{2.790} &   \multicolumn{1}{c|}{3.520} & \multicolumn{1}{c}{0.329}   \\
& \checkmark  &              & \checkmark   &   \multicolumn{1}{c|}{2.010}    &    \multicolumn{1}{c|}{2.610}   &    \multicolumn{1}{c|}{3.730}    & \multicolumn{1}{c}{0.392}  \\
 & \checkmark  &  \checkmark  & \checkmark   &  \multicolumn{1}{c|}{1.930}    &  \multicolumn{1}{c|}{2.650}    &  \multicolumn{1}{c|}{3.610}     & \multicolumn{1}{c}{0.362}   \\
\hline
\checkmark & \checkmark &  \checkmark  &              & \multicolumn{1}{c|}{1.860} & \multicolumn{1}{c|}{2.840} &  \multicolumn{1}{c|}{3.110} & \multicolumn{1}{c}{0.388}   \\
\checkmark & \checkmark &              & \checkmark   & \multicolumn{1}{c|}{1.850} & \multicolumn{1}{c|}{2.530} &  \multicolumn{1}{c|}{3.250} & \multicolumn{1}{c}{1.131}  \\
\checkmark & \checkmark & \checkmark   & \checkmark   & \multicolumn{1}{c|}{\textbf{1.660}} & \multicolumn{1}{c|}{\textbf{2.480}} &  \multicolumn{1}{c|}{\textbf{2.700}} & \multicolumn{1}{c}{\textbf{0.095}}   \\  
\hline
\end{tabular}
}
\caption{We study the performance of the architecture styles through combinations of local and global encoders and decoders on the Airplane, Car, Chair of the Shapenet dataset and KITTI dataset via Chamfer Distance metric. It can be observed that a combination of global and local encoders and decoders gives the best performance among all the possible combinations.}
\label{tbl:shapenet_ablation}
\end{table}

\begin{table}[]
\centering
\begin{tabular}{c|cccc}
\textbf{\begin{tabular}[c]{@{}c@{}}Number of regions\\ removed\end{tabular}} & \textbf{Airplane} & \textbf{Car}  & \textbf{Chair} & \textbf{KITTI} \\ \hline \hline
0                                                                            & \textbf{1.66}     & \textbf{2.48} & \textbf{2.70}  & \textbf{0.095} \\ \hline
1                                                                            & 1.67              & 2.50          & 2.73           & 0.097          \\
2                                                                            & 1.76              & 2.60          & 2.79           & 0.100          \\
3                                                                            & 1.89              & 2.67          & 2.95           & 0.102   \\      \hline
\end{tabular}
\caption{Chamfer Distance onShapenet and KITTI with varying numberof removed regions}
\label{tbl:num_region}
\end{table}

\begin{table}[]
\centering
{%
\begin{tabular}{c||cccc}
\hline
\multirow{2}{*}{\textbf{Ablation}} & \multicolumn{1}{c|}{\textbf{Airplane}} & \multicolumn{1}{c|}{\textbf{Car}} & \multicolumn{1}{c|}{\textbf{Chair}} & \textbf{KITTI}       \\ \cline{2-5} 
                                   & \multicolumn{1}{c|}{\textbf{CD}}       & \multicolumn{1}{c|}{\textbf{CD}}  & \multicolumn{1}{c|}{\textbf{CD}}    & \textbf{CD}          \\ \hline
\multicolumn{1}{l||}{\textbf{Adding third level}}                                  & \multicolumn{1}{c|}{2.070}        &     \multicolumn{1}{c|}{2.550}    &       \multicolumn{1}{c|}{3.030}      &       \multicolumn{1}{c}{0.123}    \\
\multicolumn{1}{l||}{\textbf{Our method~(2 levels)}}                                  & \multicolumn{1}{c|}{\textbf{1.660}}        &                      \multicolumn{1}{c|}{\textbf{2.480}}        &     \multicolumn{1}{c|}{\textbf{2.700}}        &    \multicolumn{1}{c}{\textbf{0.095}}                 \\
\hline
\end{tabular}
}
\caption{Quantitative Results on the architecture changes in our method. All the Chamfer Distance metric values reported for Shapenet are multiplied with 100. It can be observed that adding a third level does not give a significant improvement in the performance.}
\label{tbl:more_ablations}
\end{table}

\begin{table}[]
\centering
{
\begin{tabular}{c||cccc}
\hline
\multicolumn{1}{l||}{\textbf{Region}} & \multicolumn{1}{l}{\textbf{Airplane}} & \multicolumn{1}{l}{\textbf{Car}} & \multicolumn{1}{l}{\textbf{Chair}} & \multicolumn{1}{l}{\textbf{KITTI}} \\ \hline
\textbf{Observed Precision} &    \multicolumn{1}{c}{0.771}                                    &     \multicolumn{1}{c}{1.113}                               &             \multicolumn{1}{c}{1.767}                     &  \multicolumn{1}{c}{0.625}                                  \\
\textbf{Unobserved Precision}   &        \multicolumn{1}{c}{0.824}                                &    \multicolumn{1}{c}{1.127}                                 &     \multicolumn{1}{c}{1.844}                             &        \multicolumn{1}{c}{0.640}                            \\
\textbf{Observed Coverage}  &  \multicolumn{1}{c}{0.848}                                      &  \multicolumn{1}{c}{1.490}                                  &   \multicolumn{1}{c}{1.344}                               &     \multicolumn{1}{c}{0.531}                               \\
\textbf{Unobserved Coverage}    &       \multicolumn{1}{c}{0.857}                                 &      \multicolumn{1}{c}{1.496}                               &    \multicolumn{1}{c}{1.355}                              &   \multicolumn{1}{c}{0.648}   \\
\hline
\end{tabular}
}
\caption{Precision and Coverage for the observed and unobserved regions on the Shapenet and KITTI dataset.}
\label{tbl:obs_unobs}
\end{table}
\begin{table}[]
\centering
\begin{tabular}{c||c|c|c}
\hline
 \textbf{Method} & \textbf{Airplane} & \textbf{Car} & \textbf{Chair} \\ \hline
\textbf{DPC\cite{insafutdinov2018unsupervised}}  &       0.423        &     0.364           &  0.315            \\
\textbf{Ours} &   \textbf{0.626}                &    \textbf{0.450}            &        \textbf{0.409}  \\   
\hline
\end{tabular}
\caption{F1-Score@1\% on the Shapenet dataset.}
\label{tbl:shapenet_f1score}
\end{table}

\begin{table}[]
\centering
{
\begin{tabular}{c||cc|cccc}
\hline
\multicolumn{1}{l||}{} & \multicolumn{2}{c|}{\textbf{Airplane}}                                & \multicolumn{2}{c|}{\textbf{Cars}}                                     & \multicolumn{2}{c}{\textbf{Chairs}}                                  \\ \hline
\textbf{p}            & \multicolumn{1}{c}{\textbf{DPC \cite{insafutdinov2018unsupervised}}} & \multicolumn{1}{c|}{\textbf{Ours}} & \multicolumn{1}{c}{\textbf{DPC}} & \multicolumn{1}{c|}{\textbf{Ours}} & \multicolumn{1}{c}{\textbf{DPC}} & \multicolumn{1}{c}{\textbf{Ours}} \\ \hline
\textbf{0.4\%}       &    \multicolumn{1}{c}{0.775}                              &    \multicolumn{1}{c|}{0.775}                               &    \multicolumn{1}{c}{0.758}                              & \multicolumn{1}{c|}{0.757}              &       \multicolumn{1}{c}{0.785}                           &       \multicolumn{1}{c}{0.787}                            \\
\textbf{0.6\%}       &       \multicolumn{1}{c}{0.674}                           &    \multicolumn{1}{c|}{0.673}                                &   \multicolumn{1}{c}{0.646}                               & \multicolumn{1}{c|}{0.647}              &     \multicolumn{1}{c}{0.664}                             &   \multicolumn{1}{c}{0.664}                                \\
\textbf{0.8\%}       &     \multicolumn{1}{c}{0.490}                             &     \multicolumn{1}{c|}{0.490}                               &     \multicolumn{1}{c}{0.489}                             & \multicolumn{1}{c|}{0.489}              &  \multicolumn{1}{c}{0.498}                                &      \multicolumn{1}{c}{0.497}                             \\
\textbf{1.0\%}       &       \multicolumn{1}{c}{0.395}                           &   \multicolumn{1}{c|}{0.394}                                 &    \multicolumn{1}{c}{0.388}                              & \multicolumn{1}{c|}{0.389}              &      \multicolumn{1}{c}{0.385}                            &    \multicolumn{1}{c}{0.385}                               \\
\textbf{1.2\%}       &    \multicolumn{1}{c}{0.256}                              &  \multicolumn{1}{c|}{0.257}                                  &  \multicolumn{1}{c}{0.246}                                & \multicolumn{1}{c|}{0.245}              &     \multicolumn{1}{c}{0.265}                             &     \multicolumn{1}{c}{0.264}                              \\ \hline
\end{tabular}
}
\caption{Uniformity Metric on the Shapenet dataset compared with the baseline DPC \cite{insafutdinov2018unsupervised}.}
\label{tbl:shapenet_uniformity}
\end{table}
\begin{table}[]
\centering
\begin{tabular}{c||ll}
\hline
\textbf{p}      & \multicolumn{1}{c}{\textbf{w/o inpainting}} & \multicolumn{1}{c}{\textbf{Ours}} \\ \hline
\textbf{0.4\%} &    0.815                                &         0.750                          \\
\textbf{0.6\%} &    0.773                              &           0.658                          \\
\textbf{0.8\%} &    0.697                              &           0.527                         \\
\textbf{1.0\%} &    0.588                              &           0.496                         \\
\textbf{1.2\%} &    0.517                              &           0.384 \\
\hline
\end{tabular}
\caption{Uniformity Metric on the KITTI dataset compared with our baseline of our model without inpainting.}
\label{tbl:kitti_uniformity}
\end{table}

\begin{table}[]
\setlength{\tabcolsep}{4pt}
\begin{tabular}{p{0.28\textwidth}||p{0.001\textwidth}p{0.001\textwidth}p{0.001\textwidth}p{0.001\textwidth}p{0.001\textwidth}p{0.001\textwidth}p{0.001\textwidth}p{0.001\textwidth}}
\hline
\multicolumn{1}{l||}{\textbf{}}  & \multicolumn{1}{c}{\footnotesize{\textbf{Airplane}}} & \multicolumn{1}{c}{\footnotesize{\textbf{Cabinet}}} & \multicolumn{1}{c}{\footnotesize{\textbf{Car}}} & \multicolumn{1}{c}{\footnotesize{\textbf{Chair}}} & \multicolumn{1}{c}{\footnotesize{\textbf{Lamp}}} & \multicolumn{1}{c}{\footnotesize{\textbf{Sofa}}} & \multicolumn{1}{c}{\footnotesize{\textbf{Table}}} & \multicolumn{1}{c}{\footnotesize{\textbf{Vessel}}} \\ \hline
\textbf{\small{Ours~(w/o inpainting)}}  &   0.026                   &    0.045                            &   \multicolumn{1}{c}{0.031}                              &    0.039                            &    0.041                            &  \multicolumn{1}{l}{0.039}                               &      \multicolumn{1}{l}{0.040}                            &     \multicolumn{1}{l}{0.033}                            \\
\textbf{\small{Ours~(Self-Supervised)}} &  0.016                 &     0.027                           &    \multicolumn{1}{l}{0.024}                         &   0.027                             &  0.030                               &     \multicolumn{1}{l}{0.029}                          &    \multicolumn{1}{l}{0.025}                    &  \multicolumn{1}{l}{0.026}  \\
\hline
\textbf{\small{PCN (Fully Supervised)\cite{yuan2018pcn}}} &    0.005                 &  0.010                             &    \multicolumn{1}{c}{0.008}                            &    {0.010}                           &       {0.011}                         &   \multicolumn{1}{l}{0.011}                             &  \multicolumn{1}{l}{0.008}                              &     \multicolumn{1}{l}{0.009}                          \\
\hline
\end{tabular}
\caption{Quantitative results of comparison of our self-supervised method~(Ours) with fully supervised method, PCN~\cite{yuan2018pcn}, and ablation of our method without inpainting.}
\label{tbl:pcn_sup_self_sup}
\end{table}




\newpage

\begin{figure*}[hbt!]
 	\centering
 	\includegraphics[width=0.99\textwidth]{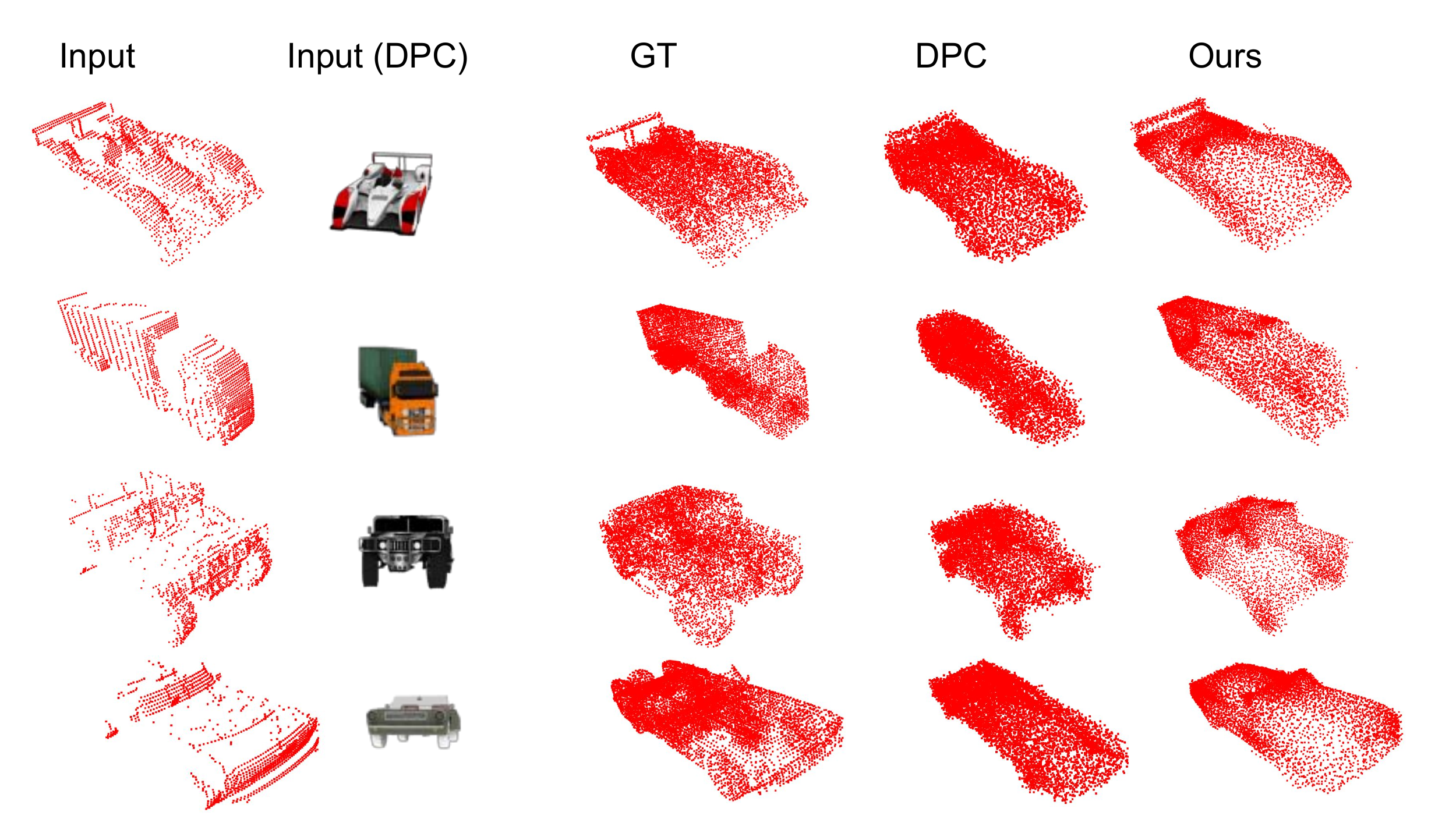}
 	\caption{Failure cases on the Car category of the ShapeNet dataset. We compare our method with the results of DPC~\cite{insafutdinov2018unsupervised}. Our method fails to create the detailed shapes of various sports cars.  Further, for the truck in the second row, our method fails to create a gap between the front and back of a truck.}       
 	\label{fig:shapenet_car_fail}
\end{figure*} 

\begin{figure*}[hbt!]
 	\centering
 	\includegraphics[width=0.99\textwidth]{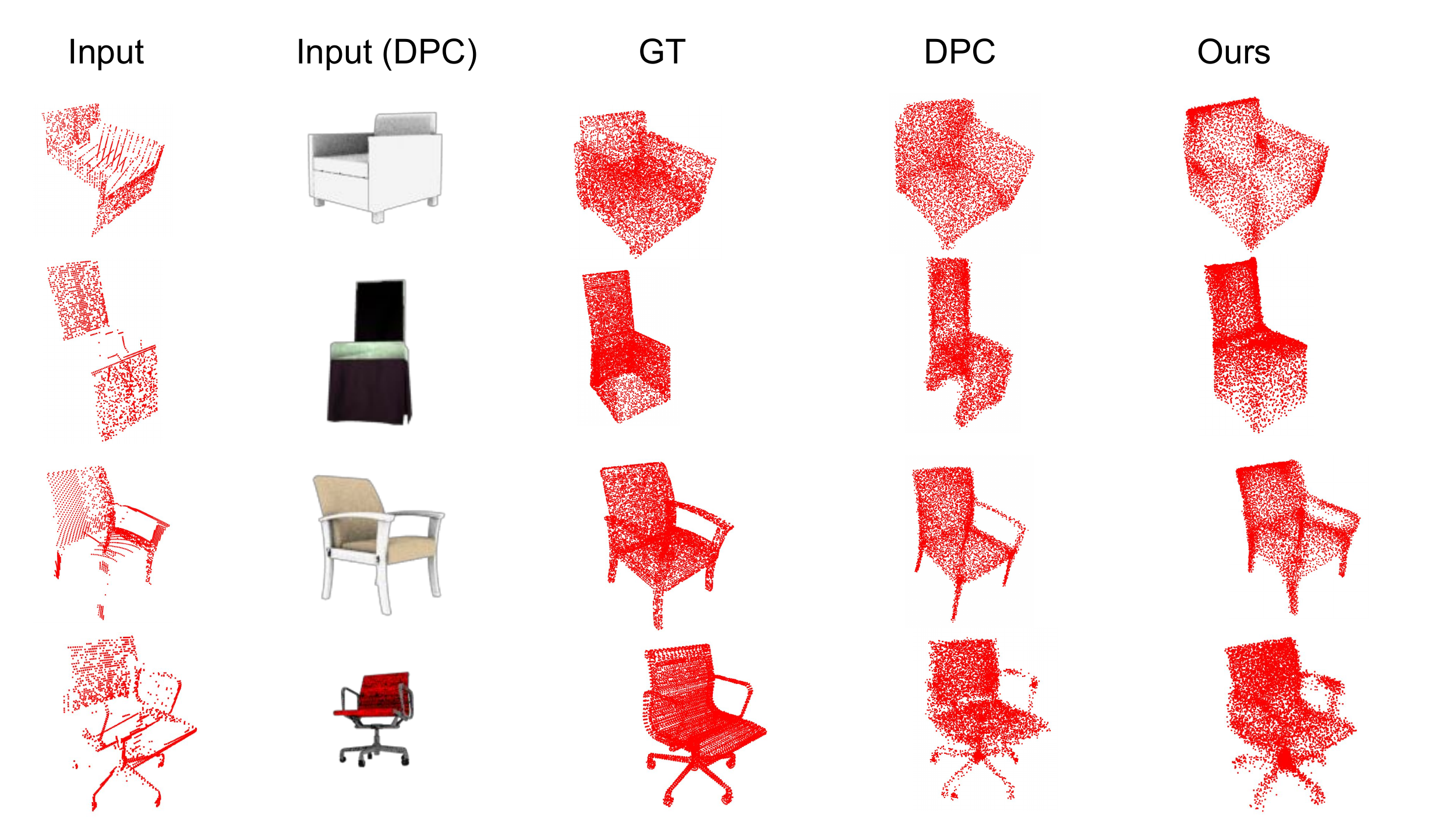}
 	\caption{Success cases on the Chair category of the ShapeNet dataset. We compare our method with the results of DPC~\cite{insafutdinov2018unsupervised}. Our method is able to show finer completion results on different types of chairs such as a sofa and desk chair than DPC~\cite{insafutdinov2018unsupervised}. It is able to complete the front, back, and arms of the chair.}       
 	\label{fig:shapenet_chair_success}
\end{figure*} 

\vspace{2 cm}

\begin{figure*}[hbt!]
 	\centering
 	\includegraphics[width=0.99\textwidth]{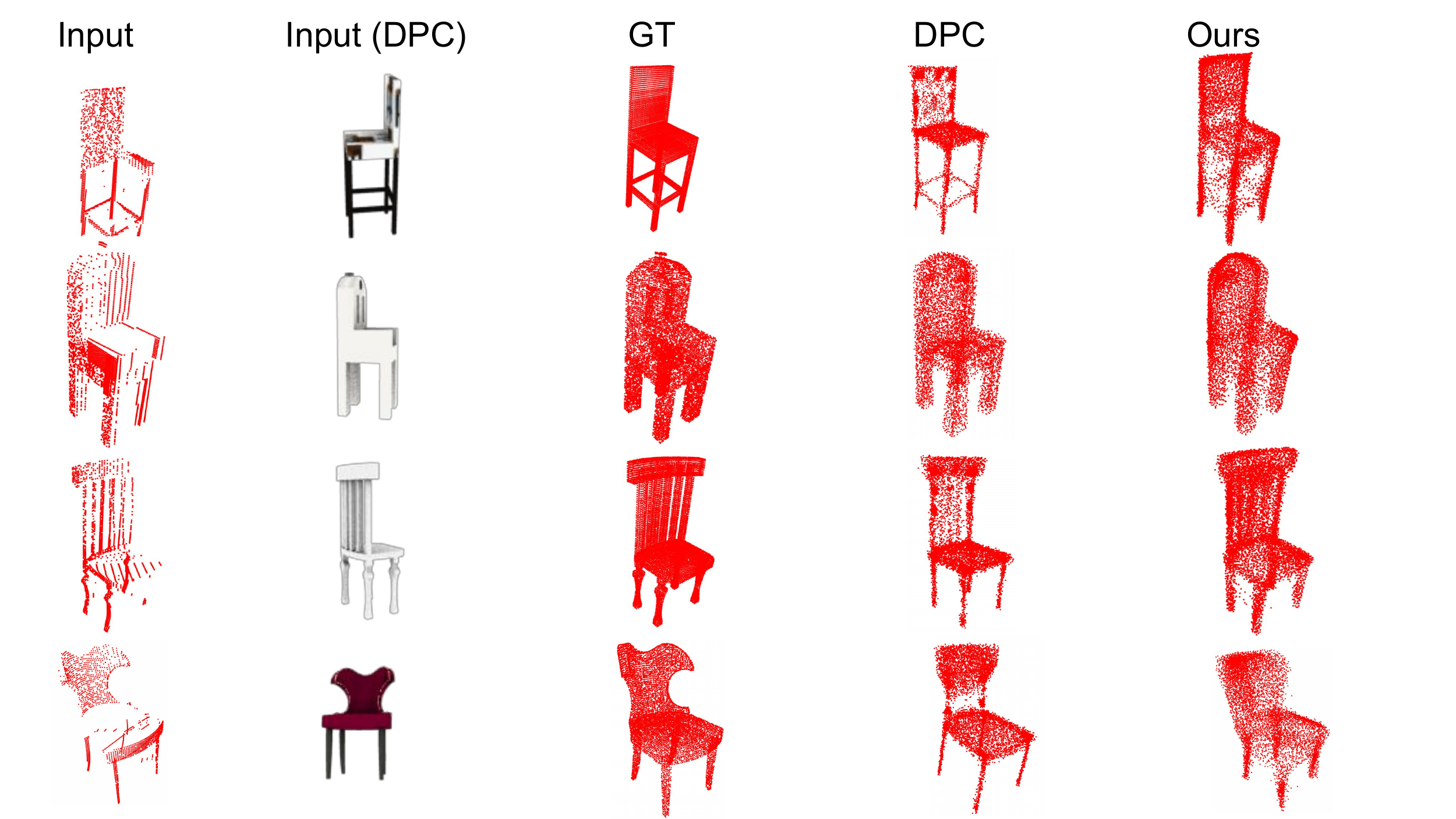}
 	\caption{Failure cases on the Chair category of the ShapeNet dataset. We compare our method with the results of DPC~\cite{insafutdinov2018unsupervised}. It can be observed in these cases that the network generates noisy results especially near the legs of a chair.}       
 	\label{fig:shapenet_chair_fail}
\end{figure*} 

\vspace{2 cm}

\begin{figure*}[hbt!]
\centering
 \includegraphics[width=0.99\textwidth]{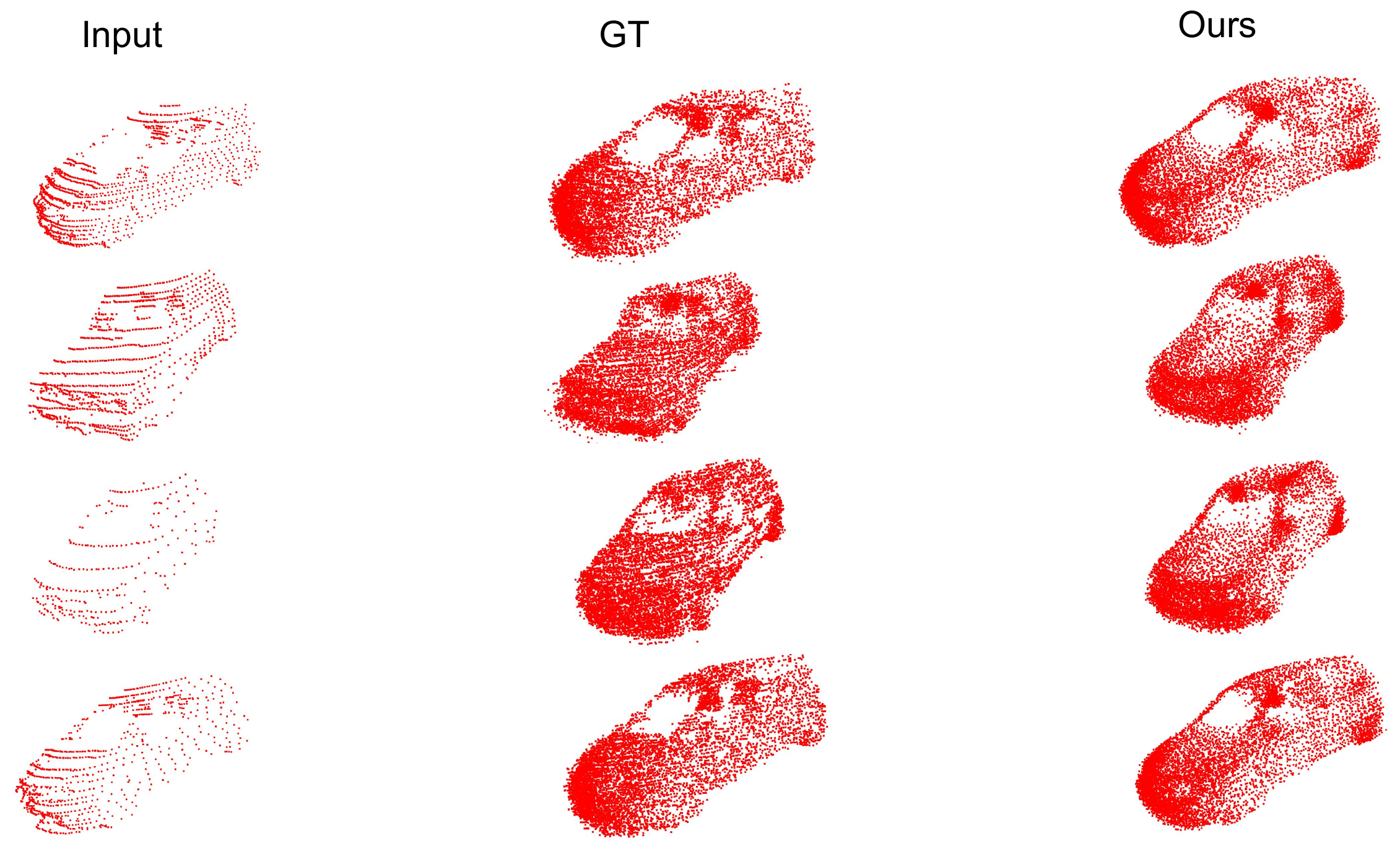}
 \caption{Completion of partial point cloud cars from the LiDAR scans of the Semantic KITTI dataset (first and second row). The third and fourth row show some failure cases of the network where the network is unable to generate the smaller details such as a tire of a car.}       
 \label{fig:kitti_supp}
\end{figure*} 

\begin{figure*}[hbt!]
\centering
 \includegraphics[width=0.99\textwidth]{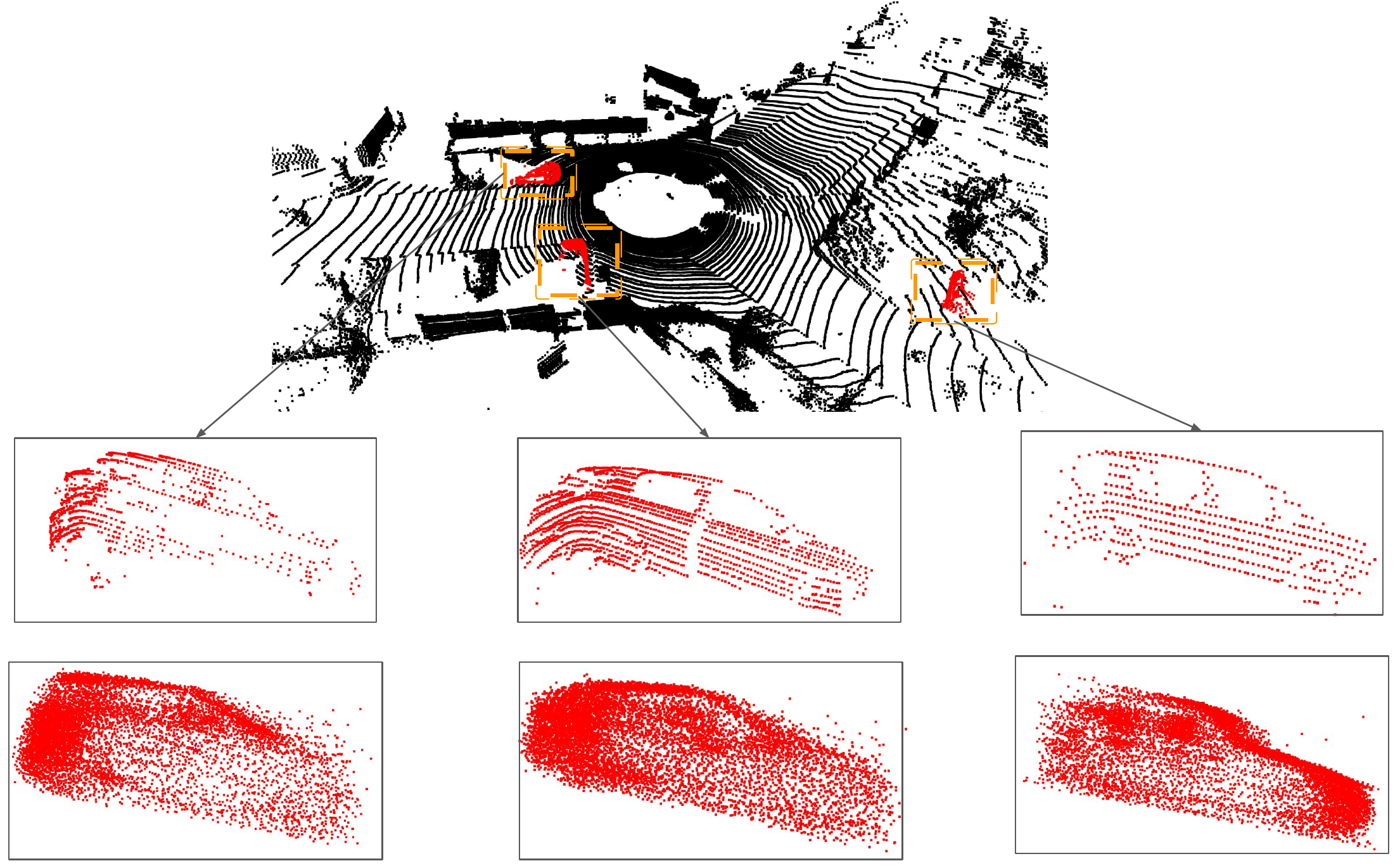}
 \caption{Completion of partial point cloud of cars in a LiDAR scan of the Semantic KITTI dataset.}       
 \label{fig:kitti_scene}
\end{figure*} 

\end{document}